\documentclass[runningheads]{llncs}

\usepackage{eccv}

\usepackage{eccvabbrv}

\usepackage{graphicx}
\usepackage{booktabs}
\usepackage{comment}
\usepackage[accsupp]{axessibility}  %
\usepackage{multirow}
\usepackage{siunitx}
\usepackage[table]{xcolor}
\usepackage{wrapfig}
\usepackage{titletoc}
\usepackage{bibunits}

\definecolor{lightgreen}{rgb}{0.9, 1.0, 0.9}\newcommand{\up}{\raisebox{0.1ex}{\scriptsize$\uparrow$}}
\newcommand{\down}{\raisebox{0.1ex}{\scriptsize$\downarrow$}}

\usepackage{hyperref}

\usepackage{orcidlink}
\usepackage[shortcuts]{extdash}

\begin{document}

\title{From Local Geometry to Global Pseudo-Labeling for Robust Positive–Unlabeled Learning under Covariate Shift}

\titlerunning{S-PUNA for Robust PU Learning under Covariate Shift}

\author{Firas Gabetni\inst{1,2} \and
Alexandre Rocchi\inst{1,2} \and
Nacim Belkhir\inst{1} \and 
Ziyi Liu\inst{1} \and 
Gianni Franchi\inst{2}
}

\authorrunning{F.~Gabetni et al.}

\institute{U2IS, ENSTA,
Institut Polytechnique de Paris
\and
AMIAD, Pôle Recherche, Palaiseau}

\maketitle
\begin{abstract}
Detecting covariate shift is critical for building reliable vision systems. While most prior work focuses on improving robustness to shift, explicitly detecting covariate shift remains underexplored. Existing approaches typically rely on fully supervised training, requiring labeled examples from both original and shifted distributions, which is often impractical. In this paper, we show that covariate shift detection can be effectively addressed with weaker supervision using Positive–Unlabeled (PU) learning. However, under covariate shift, in-distribution and shifted data overlap significantly, making classical PU methods unstable and sensitive to noise. To overcome this challenge, we introduce \textbf{Spectral PU Neighborhood Annotation (S-PUNA)}, a geometry-aware framework that progressively discovers shifted data by leveraging the local manifold structure of visual features. Extensive experiments show that \textbf{S-PUNA} achieves state-of-the-art performance in PU settings and remarkably matches the performance of fully supervised methods. Moreover, our approach transfers robustly across different types of shifts, demonstrating strong generalization capabilities. Code is available at \url{https://github.com/fira7s/S-PUNA}.
\end{abstract}

\section{Introduction}
\label{sec:intro}
Modern Deep Neural Networks (DNNs) are usually trained on a source distribution and tested on data assumed to follow a similar distribution. A central objective of classical machine learning research has been to improve generalization under mild distributional shifts, commonly referred to as \emph{covariate shift}, where the input distribution $P_X$ changes but $P(Y \mid X)$ stays the same. Many works therefore focus on domain generalization \cite{zhou2022domain}, domain adaptation \cite{ajith2023domain}, and robustness to small input perturbations \cite{hendrycksbenchmarking}. 
In contrast, under a \emph{semantic shift} \cite{yang2024generalized} (for example when a model trained on animals is tested on objects) the predictions become unreliable. In such cases, the goal is not to generalize but to detect the shift and raise an alert. Thus, models should be invariant to covariate shift but sensitive to semantic shift.

In this work, we argue that \emph{detecting covariate shift itself} is also an important problem that has been less studied. In many real applications, knowing that the data distribution has changed is critical. For example, detecting AI-generated data often relies on identifying small distribution differences \cite{zhu2023gendet, yu2021survey}. Similarly, in finance \cite{dixon2020machine}, healthcare \cite{finlayson2021clinician}, or sensor systems \cite{Lu2019LearningUC}, gradual covariate shifts can signal that model predictions may become unreliable. 

Existing approaches are predominantly fully supervised. For example, recent work such as DisCoPatch \cite{caetano2025discopatch} trains a dedicated discriminator to distinguish between real and synthetic images. While effective, such methods require labeled examples from both distributions, limiting their applicability when negative samples are unavailable or expensive to curate. This raises a fundamental question:

\begin{center}
\emph{Are we forced to rely on fully supervised training for covariate shift detection, or can we relax the level of supervision?}
\end{center}

We propose to address covariate shift detection through the lens of \emph{Positive–Unlabeled (PU) learning}. This field studies binary classification when only labeled positive examples and unlabeled data are available. This paradigm naturally arises in anomaly detection \cite{takahashi2024deep}, medical diagnosis \cite{bekker2020learning}, or out-of-distribution detection \cite{katz2022training}, where negative labels are scarce or unreliable. Surprisingly, to the best of our knowledge, PU learning has not been systematically investigated for covariate shift detection.

However, directly applying classical PU risk minimization in the presence of covariate shift is problematic. Standard PU methods rely on unbiased risk estimators that assume sufficient separability between positive and negative distributions. Under covariate shift, the distributions may overlap significantly, leading to a small total variation distance and high intrinsic Bayes risk. In this regime, global risk estimation becomes unstable and prone to high variance.

We argue that in covariate-shifted PU settings, the negative distribution should not be estimated globally. Instead, it should be \emph{progressively discovered} through its local geometric structure. Motivated by the manifold hypothesis, we treat the unlabeled dataset as a mixture of low-dimensional submanifolds corresponding to positive and shifted negative components. Rather than directly estimating risk differences, we iteratively uncover the negative manifold by propagating reliable local neighborhood information from positive anchor points.

\begin{figure}[!h]
    \centering
    \vspace{-20pt}
    \includegraphics[width=\linewidth]{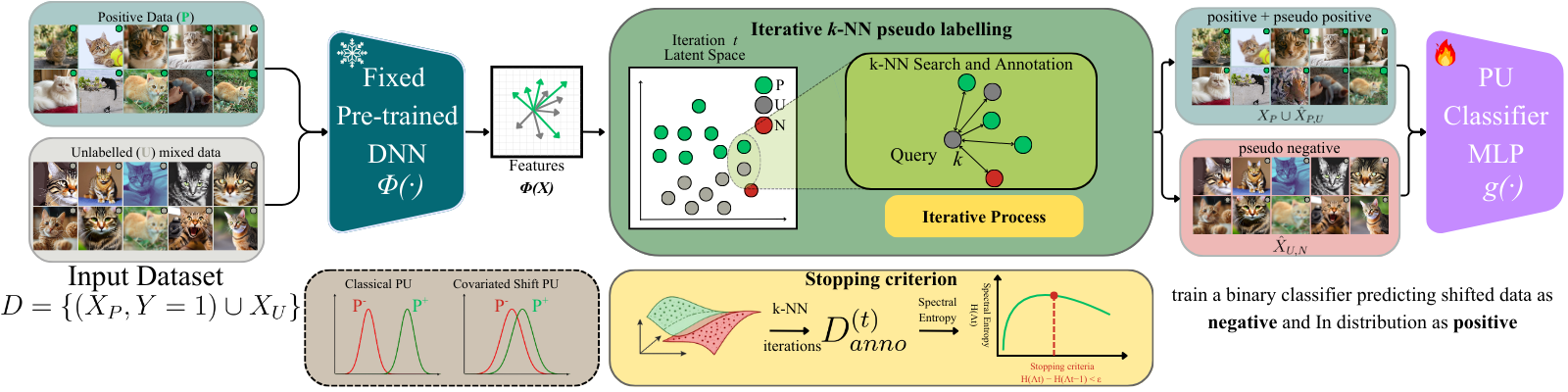}
    \vspace{-10pt}
    \caption{\textbf{Overview of \textbf{S-PUNA}.} Given a PU dataset $D$, we use a pre-trained DNN $\Phi(\cdot)$ that maps samples to the latent feature space $\Phi(X)$. \textbf{S-PUNA} then performs \emph{iterative $k$-NN pseudo-labeling}. $k$-NN expands both pseudo-positive $\tilde{X}_{P,U}$ and pseudo-negative $\tilde{X}_{U,N}$ samples symmetrically. The process is stopped using the spectral-entropy criterion. The PU classifier $g(\cdot)$ is trained to separate positives and negatives.}
    \label{fig:overview_s-puna}
    \vspace{-20pt}
\end{figure}

To operationalize this idea, we introduce \textbf{Spectral PU Neighborhood Annotation (S-PUNA)}, a geometry-aware pseudo-labeling framework for covariate shift detection as illustrated in Figure \ref{fig:overview_s-puna}. \textbf{S-PUNA} initializes from labeled positives and iteratively expands both positive and negative anchor sets using confidence-controlled $k$-nearest neighbor propagation in feature space. Unlike classical $k$-NN-PU \cite{zhang2009reliable} that propagates only positive labels, \textbf{S-PUNA} symmetrically constructs both classes, progressively refining the decision boundary.

A key challenge in iterative pseudo-labeling is preventing semantic drift. We address this through a novel \emph{spectral entropy stopping criterion}. By monitoring the eigenvalue distribution of the covariance matrix of the discovered negative set, we measure its intrinsic dimensionality and structural complexity. The iterative expansion stops once the spectral entropy stabilizes, indicating that the negative manifold has been sufficiently captured without invading overlapping regions.

Finally, we train a DNN classifier to distill the non-parametric neighborhood structure to maximize latent separation and distinguish positive from shifted components.

Our contributions are summarized as follows:
\begin{itemize}
    \item We are the first to use PU for detecting changes in covariates and to provide theoretical insights on this new problem.
    \item We introduce \textbf{S-PUNA}, a geometry-aware framework that progressively uncovers the shifted distribution through local manifold expansion.
    \item We propose a principled spectral entropy stopping criterion that detects early contamination and prevents pseudo-labeling drift under covariate overlap.
    \item %
    We provide theoretical insights into why spectral entropy signals help prevent annotation contamination.
    \item We build new covariate shift benchmarks and demonstrate state-of-the-art performance, approaching fully supervised methods, while showing strong transferability across diverse shifts.
\end{itemize}

\section{Related Works}

\textbf{Domain generalization}. 
 Domain adaptation (DA)\cite{ajith2023domain} and domain generalization (DG)\cite{zhou2022domain} also study changes in distribution, but their objective is fundamentally different from ours.  In DA/DG, covariate shift is usually treated as a nuisance: the goal is to learn representations or classifiers that generalise and remain accurate and invariant across domains. In contrast, our goal is not to adapt to the shifted distribution or to improve task accuracy in the event of a shift; rather, it is the shift itself that is the target to be detected. This distinction is important because methods designed for DA/DG may deliberately remove or obscure domain-specific information, whereas our method must preserve and exploit this information to detect the presence of a shift.

\textbf{Covariate Shift Detection.} %
While the problem is often included within broader discussions of general OOD detection \cite{yang2024generalized}, isolating and detecting these non-semantic shifts remains an under-explored challenge. Fully supervised strategies like DiscoPatch \cite{caetano2025discopatch} require the unrealistic assumption that the specific covariate shifts are represented in the training set. Meanwhile, standard OOD algorithms like \cite{ammar2023neco} and spatial anomaly detection methods like PatchCore \cite{roth2022towards} and SimpleNet \cite{liu2023simplenet} excel in semantic or localized tasks but prove ineffective for image-level covariate shift detection (see Appendix \ref{appendix:ood_detection_covariateshift} for detailed results).

\textbf{Positive-Unlabeled (PU) learning.} PU learning enables binary classification using only labeled positives and an unlabeled mixture \cite{bekker2020learning}. Modern approaches rely on disambiguation-free empirical risk estimators. For instance, nnPU \cite{kiryo2017nnpu} prevents overfitting by enforcing non-negative risk constraints, DCPU \cite{Li2025DCPU} handles biased or disjoint positive sets, and DistPU \cite{zhao2022distpu} mitigates negative-prediction bias by aligning predictions with the class prior through entropy minimization. PU formulations have also been applied to open-world OOD detection by treating nominal data as positives and deployment data as an unlabeled mixture \cite{blanchard2010semi, pang2019deep}.
Most existing methods assume a semantic shift between the P and N, which simplifies the PU framework. While some works \cite{sakai2019covariate,kumagai2025positive,chan2020unlabelled} address PU under covariate shift, they focus on classifier generalization between train and test sets rather than using the PU framework for shift detection.

\section{Preliminaries and mathematical Insights}\label{sec:Preliminaries}

\subsection{Distribution Shift Detection}

\begin{figure}[!h]
    \centering
    \vspace{-20pt}
    \includegraphics[width=\linewidth]{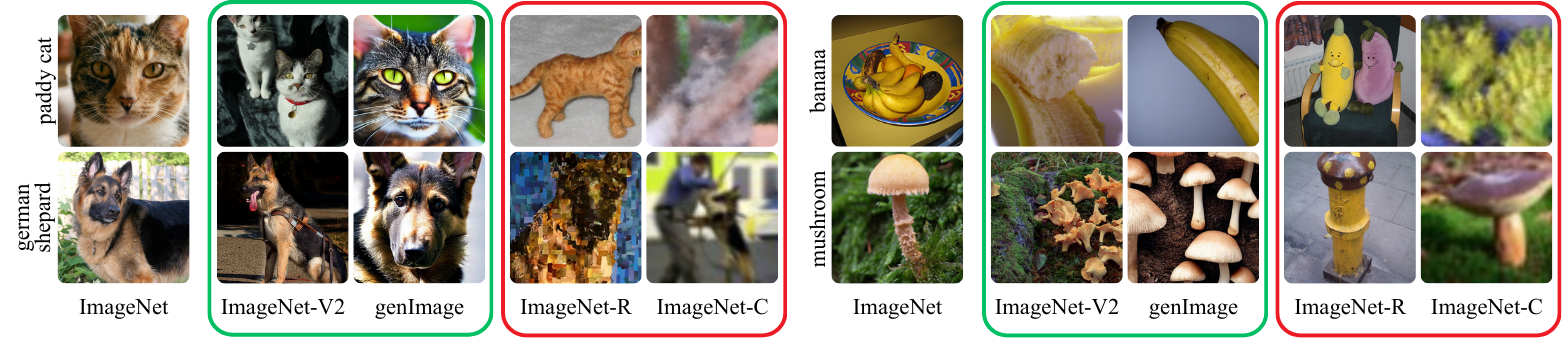}
    \vspace{-10pt}
    \caption{\textbf{Covariate shifts in ImageNet benchmarks. }Comparison of natural images (ImageNet, ImageNet-V2) against synthetic corruptions (ImageNet-C), stylized rendering (ImageNet-R), and generative data (GenImage) for the classes Paddy cat (Near shifts are represented in green while Far shifts are in red).}
    \label{fig:covariate_shift_examples}
    \vspace{-20pt}
\end{figure}

\textit{Notations and preliminaries.}
Let $\mathcal{X}$ be the set of all possible inputs  
(usually points in $\mathbb{R}^d$, i.e. $d$-dimensional vectors). Let $\mathcal{Y} = \{1, 2, \dots, K\}$ be the set of possible class labels 
(there are $K$ different classes). During training we receive a dataset
\(
\mathcal{D}_{\mathrm{train}} = \{(x_1,y_1), (x_2,y_2), \dots, (x_n,y_n)\} \)
where each pair $(x_i, y_i)$ is supposed to come from the same joint distribution
$P_{\mathrm{train}}(X,Y)$. We train a classifier $f_\theta$ by minimizing the average training loss:
\begin{equation}
\hat{\theta} = \arg\min_{\theta} \ 
\frac{1}{n} \sum_{i=1}^n \ell\bigl( f_\theta(x_i),\ y_i \bigr)
\end{equation}

Here $\ell(\cdot,\cdot)$ denotes a loss function. Classical learning theory assumes that future test samples $(X,Y)$ follow the same distribution:
\(
P_{\mathrm{test}}(X,Y) = P_{\mathrm{train}}(X,Y).
\)
In practice this assumption rarely holds, leading to \textbf{distribution shift}. Detecting and handling such shifts is essential for reliable machine learning systems. Two main types are commonly distinguished: \textbf{Semantic shift} and \textbf{Covariate shift}.

\textit{Semantic Shift} occurs when the label distribution differs between training and test data, i.e.,
\(
P_{\mathrm{test}}(Y) \neq P_{\mathrm{train}}(Y).
\)
This setting is closely related to \textit{Out-of-Distribution (OOD) detection}, where test samples are drawn from a distribution $P_{\mathrm{test}}$ defined on the same input space $\mathcal{X}$ but involving unseen semantic classes. The OOD literature typically distinguishes two regimes \cite{yang2022openood}: \textbf{Far OOD} where the new classes are semantically very different from those seen during training, and \textbf{Near OOD} where they remain semantically similar. Near-OOD detection is generally more challenging due to the stronger semantic overlap with the training classes.

\textit{Covariate Shift} refers to a change in the input distribution while the conditional labeling function remains unchanged:
\(
P_{\mathrm{test}}(X) \neq P_{\mathrm{train}}(X),
\qquad
P_{\mathrm{test}}(Y \mid X) = P_{\mathrm{train}}(Y \mid X).
\)
In this case, the optimal classifier is theoretically unchanged, but empirical risk minimization may become biased since the training data no longer reflect the marginal distribution encountered at test time. As with semantic shift, one can distinguish between \textbf{Near Shift}, where $P_{\mathrm{test}}(X)$ differs only slightly from $P_{\mathrm{train}}(X)$, and \textbf{Far Shift} corresponding to larger changes in input statistics. This distinction primarily reflects the difficulty of the shift rather than a strict quantitative measure in $\mathcal{X}$.

\subsection{Background and Mathematical Insights}\label{sec:Background}
\paragraph{Classical Binary Classification.}
Let $(X,Y) \sim P$ be a random pair taking values in $\mathcal{X}\times\{-1,+1\}$,
where $P$ denotes the joint data-generating distribution.
We denote by
\(\pi_p = \mathbb{P}(Y=+1)\),
and 
\(
\pi_n = \mathbb{P}(Y=-1)=1-\pi_p
\)
, the class priors, and by
\(
P^+ = P_{X|Y=+1}\) and 
\(
P^- = P_{X|Y=-1}
\)
, the class-conditional distributions.

Given a hypothesis class $\mathcal{H}$ of measurable functions
$f:\mathcal{X}\to\mathbb{R}$ and a loss function
$\ell:\mathbb{R}\times\{-1,+1\}\to[0,1]$,
the \emph{population (true) risk} of a classifier $f$ is defined as
\begin{equation}
R(f)
=
\mathbb{E}_{(X,Y)\sim P}[\ell(f(X),Y)].
\end{equation}
Using the law of total expectation, this can be decomposed as
\begin{equation}
R(f)
=
\pi_p \, \mathbb{E}_{X\sim P^+}[\ell(f(X),+1)]
+
\pi_n \, \mathbb{E}_{X\sim P^-}[\ell(f(X),-1)].
\label{eq:true-risk-decomposition}
\end{equation}
In practice, $P$ is unknown and we observe an \emph{i.i.d.}\ training sample such that
\(
D=\{(x_i,y_i)\}_{i=1}^n \sim P^n.
\)
\newline The \emph{empirical risk} is defined as
\begin{equation}
\hat R_n(f)
=
\frac1n \sum_{i=1}^n \ell(f(x_i),y_i).
\end{equation}
Empirical Risk Minimization (ERM) selects
\(
\hat f \in \arg\min_{f\in\mathcal{H}} \hat R_n(f).
\)

\paragraph{Positive–Unlabeled (PU) Classification}

In PU learning, we do not observe labeled negative examples.
Instead, we observe: a labeled positive set
    $\mathcal{X}_P = \{x_i^p\}_{i=1}^{n_p} \sim (P^+)^{n_p}$, and an unlabeled set
    $\mathcal{X}_U = \{x_j^u\}_{j=1}^{n_u} \sim (P_X)^{n_u}$, where the marginal distribution satisfies
\begin{equation}
P_X = \pi_p P^+ + \pi_n P^-.
\label{eq:marginal-mixture}
\end{equation}
Similarly to \cite{du2014analysis} a key observation is that the negative component of the risk can be
recovered from the mixture structure.
Indeed, by linearity of expectation and~\eqref{eq:marginal-mixture},
\begin{align}
\mathbb{E}_{X\sim P_X}[\ell(f(X),-1)]
&=
\pi_p \mathbb{E}_{X\sim P^+}[\ell(f(X),-1)]
+
\pi_n \mathbb{E}_{X\sim P^-}[\ell(f(X),-1)].
\end{align}
Rearranging yields:
\begin{equation}
\pi_n \mathbb{E}_{X\sim P^-}[\ell(f(X),-1)]
=
\mathbb{E}_{X\sim P_X}[\ell(f(X),-1)]
-
\pi_p \mathbb{E}_{X\sim P^+}[\ell(f(X),-1)].
\label{eq:negative-recovery}
\end{equation}
Substituting~\eqref{eq:negative-recovery} into~\eqref{eq:true-risk-decomposition}
gives the \emph{unbiased PU risk}:
\begin{equation}
R_{pu}(f)
=
\pi_p \mathbb{E}_{X\sim P^+}
\big[
\ell(f(X),+1)-\ell(f(X),-1)
\big]
+
\mathbb{E}_{X\sim P_X}
[\ell(f(X),-1)].
\label{eq:pu-risk}
\end{equation}
Importantly, $R_{pu}(f)=R(f)$ for all $f$,
hence minimizing $R_{pu}$ is equivalent to minimizing the true risk.

An empirical estimator of~\eqref{eq:pu-risk} is
\begin{equation}
\hat R_{pu}(f)
=
\frac{\pi_p}{n_p}
\sum_{i=1}^{n_p}
\tilde\ell(f(x_i^p))
+
\frac{1}{n_u}
\sum_{j=1}^{n_u}
\ell(f(x_j^u),-1),\text{ where}
\end{equation}
\begin{equation}
\tilde\ell(f(x))
=
\ell(f(x),+1)-\ell(f(x),-1).
\end{equation}

\paragraph{Rademacher Generalization Bound}

Let $\mathcal{H}$ be a class of functions taking values in $[0,1]$.
We can define its empirical Rademacher complexity:
\begin{equation}
\mathcal R_n(\mathcal H)
=
\mathbb{E}_\sigma
\left[
\sup_{f\in\mathcal H}
\frac1n\sum_{i=1}^n
\sigma_i f(x_i)
\right],\text{ where $\sigma_i$ are i.i.d.\ Rademacher variables.}
\end{equation}

A standard uniform convergence result (see, e.g.,
 Theorem 3.1 of \cite{mohri2018foundations})
states that for any $\delta>0$, with probability at least $1-\delta$,
\begin{equation}
\sup_{f\in\mathcal H}
|R(f)-\hat R_n(f)|
\le
2\mathcal R_n(\mathcal H)
+
\sqrt{\frac{\log(1/\delta)}{2n}}.
\label{eq:rad-bound}
\end{equation}
This inequality provides a distribution-dependent control of the generalization gap.

Applying~\eqref{eq:rad-bound} separately to the positive
and unlabeled samples yields, with probability at least $1-\delta$,
\begin{align}
R_{pu}(f)
\le
\hat R_{pu}(f)
&+
2\pi_p \mathcal R_{n_p}(\mathcal H)
+
2\mathcal R_{n_u}(\mathcal H)
\nonumber\\
&+
\pi_p \sqrt{\frac{\log(1/\delta)}{2n_p}}
+
\sqrt{\frac{\log(1/\delta)}{2n_u}}.
\label{eq:pu-generalization}
\end{align}
Thus, the excess risk is controlled by the complexity
of the hypothesis class and the sample sizes
$n_p$ and $n_u$.

\paragraph{Bayes Optimal Risk and Total Variation}

Assume now balanced classes:
\begin{equation}    
\pi_p=\pi_n=\frac12.
\end{equation}
Define the regression function
\(
\eta(x)=\mathbb{P}(Y=+1\mid X=x).
\)

We can define the Bayes classifier, as in \cite{bishop2006pattern}, by
\(    
f^*(x)=\mathbf 1\{\eta(x)\ge 1/2\},
\) which minimizes $R(f)$ over all measurable classifiers.

Its risk is
\begin{equation}
R(f^*)
=
\mathbb{E}[\min(\eta(X),1-\eta(X))].
\end{equation}
Using the identity
\begin{equation}
\min(a,b)=\frac12(a+b-|a-b|),
\end{equation}
and the fact that
\begin{equation}
\|P^+-P^-\|_{TV}
=
\frac12\int |p^+(x)-p^-(x)|dx,
\end{equation}
one obtains the classical result from statistical decision theory:
\begin{equation}
R(f^*)
=
\frac12
\left(
1-\|P^+-P^-\|_{TV}
\right).
\label{eq:bayes-tv}
\end{equation}

Equation~\eqref{eq:bayes-tv} shows that the intrinsic difficulty
of the classification problem is entirely governed by the
total variation distance between the class-conditional distributions.
If the distributions overlap significantly,
$\|P^+-P^-\|_{TV}$ is small and the Bayes error is large.
If they are well separated,
$\|P^+-P^-\|_{TV}$ approaches $1$ and the Bayes error approaches $0$.

\subsection{Implications for Covariate Shift Detection}

Under covariate shift,
the marginal distribution of $X$ changes while
$P(Y\mid X)$ remains fixed.
When $P^+$ and $P^-$ differ substantially,
the total variation term in~\eqref{eq:bayes-tv}
captures this shift directly.

Combining~\eqref{eq:pu-generalization}
with~\eqref{eq:bayes-tv},
we obtain the high-probability bound
\begin{align}
R_{pu}(\hat{f})
\le
\frac12
\left(
1-\|P^+-P^-\|_{TV}
\right)
+
2\pi_p \mathcal R_{n_p}(\mathcal H)
+
2\mathcal R_{n_u}(\mathcal H)
\nonumber\\
+
\pi_p \sqrt{\frac{\log(1/\delta)}{2n_p}}
+
\sqrt{\frac{\log(1/\delta)}{2n_u}}.
\end{align}
This inequality highlights two sources of difficulty: \textbf{Intrinsic difficulty}, governed by
    $\|P^+-P^-\|_{TV}$,
    which reflects distributional overlap and the \textbf{estimation error}, governed by
    Rademacher complexity and sample size.

In extreme classical shift scenarios used in PU, where
$P^+$ and $P^-$ have nearly disjoint supports,
$\|P^+-P^-\|_{TV}\approx 1$,
and the intrinsic Bayes error is small. However, in our case $\|P^+-P^-\|_{TV}\approx 0$, hence the covariate shift might complicate all the training.

\section{Spectral-PU Neighborhood Annotation (S-PUNA)}

In the covariate shift regime, the class-conditional distributions
$P^+$ and $P^-$ may exhibit substantial overlap in the ambient space,
leading to a small total variation distance
$\|P^+ - P^-\|_{TV}$.
As shown in Section~\ref{sec:Background},
the Bayes risk scales as
$
R(f^*) = \tfrac12(1-\|P^+ - P^-\|_{TV}),
$
implying that when $\|P^+ - P^-\|_{TV}$ is small,
the intrinsic classification problem is difficult.

To mitigate this effect, we propose
\textbf{Spectral--PU Neighborhood Annotation (S-PUNA)},
a progressive, geometry-aware pseudo-labeling framework.
Rather than estimating the negative distribution globally,
\textbf{S-PUNA} exploits the \emph{local manifold structure}
of the unlabeled data to iteratively uncover
regions that are statistically inconsistent
with the positive anchor distribution.
This results in a gradual expansion of a reliable
negative support estimate.
The Appendix \ref{appendix:theoretical_justification} gives a deeper theoretical overview of the soundness of our proposed approach.

\subsection{Iterative $k$-NN-Based Pseudo-Labeling}
Our method is a two-stage process as illustrated in Figure \ref{fig:overview_s-puna}: an initialization phase of the $k$-NN, followed by symmetric manifold expansion.

Let us consider that we have two sets of pseudo-annotations, one for positive and one for unlabeled negative data, which we denote as:
\(
\hat{\mathcal{X}}_{U,P}^{(t_0)} = \emptyset,
\qquad
\hat{\mathcal{X}}_{U,N}^{(t_0)} = \emptyset
\)  for the first iteration $t_0$.
Then for the next iterations $t=t_0+1,\dots,T$,
we maintain the current annotated sets:
\begin{equation}
\mathcal{X}_P^{(t)}
=
\mathcal{X}_P
\cup
\hat{\mathcal{X}}_{U,P}^{(t-1)},
\qquad
\mathcal{X}_N^{(t)}
=
\hat{\mathcal{X}}_{U,N}^{(t-1)}.
\end{equation}

\paragraph{Baseline and Sets Initialization.}
All samples features $\phi(x)$ are extracted from the sixth block of a pre-trained Vision Transformer (ViT) \cite{dosovitskiyimage}. 
This specific intermediate representation is chosen for its efficacy in capturing the geometric nuances of covariate shifts (see Appendix \ref{appendix:vit_layer_selection_covariate_shifts} for a comparative analysis of feature layer sensitivity). 
Our technique starts by constructing a $k$-nearest neighbor ($k$-NN) index on Positive Domain (PD) training features, $\mathcal{X}_{train}$. For each sample $x$ in the unlabeled pool $\mathcal{X}_U$, we compute an initial distance score $s_{orig}(x)$ as the minimum distance from the sample to its $k$-nearest neighbors in $\mathcal{X}_{train}$. We then initialize the positive (PD) and negative (covariate) sets for the first iteration $t_0$ by ranking the pool according to these scores:
\begin{equation}
    \hat{\mathcal{X}}_{U,c}^{(t_0)}=\begin{cases}\{x\in\mathcal{X}_U\mid x\text{ is among top }\alpha\text{ lowest }s_{orig}(\cdot)\},&\text{if }c=P\\\{x\in\mathcal{X}_U\mid x\text{ is among top }\alpha\text{ highest }s_{orig}(\cdot)\},&\text{if }c=N\end{cases}
\end{equation}
where $\alpha$ represents a small number of features used to seed the initial pseudo-labeled sets. For subsequent iterations $t=t_0+1, \dots, T$, the current annotated sets are maintained as:
\(
\mathcal{X}_P^{(t)} = \mathcal{X}_P \cup \hat{\mathcal{X}}_{U,P}^{(t-1)}, \qquad \mathcal{X}_N^{(t)} = \hat{\mathcal{X}}_{U,N}^{(t-1)} \qquad \mbox{and } \mathcal{X}_U^{(t)} = \mathcal{X}_U^{(t-1)} / (\hat{\mathcal{X}}_{U,N}^{(t)}  \cap \hat{\mathcal{X}}_{U,P}^{(t)} ).
\)
\paragraph{Neighborhood Search and Scoring.}
At each iteration $t$, we compute two separate distance metrics for each candidate sample $x \in \mathcal{X}_U^{(t)}$:
\begin{enumerate}
    \item $d_{P}(x)$: the mean distance from the sample features  $\phi(x)$ to its $k$-nearest neighbors in the positive bank $\mathcal{X}_P^{(t)}$.
    \item $d_{N}(x)$: the mean distance from the sample features  $\phi(x)$ to its $k$-nearest neighbors in the negative bank $\mathcal{X}_N^{(t)}$.
\end{enumerate}
Using these distances we calculate a score for ranking:
\begin{equation}
s_t(x) = d_{ood}(x) - d_{id}(x).
\end{equation}

\paragraph{Set Expansion.}
The set is progressively expanded by ranking the candidates based on $s_t(x)$. We select the top $\beta$ samples for each class to be added to the respective sets:
\begin{equation}
\hat{\mathcal{X}}_{U,c}^{(t)}=\hat{\mathcal{X}}_{U,c}^{(t-1)}\cup\begin{cases}\{x\mid x\text{ has top }\beta\text{ highest }s_t(x)\},&\text{if }c=P\\\{x\mid x\text{ has top }\beta\text{ lowest }s_t(x)\}.&\text{if }c=N\end{cases}
\end{equation}
$\beta$ is a fixed parameter we set during the whole algorithm that controls the number of samples added to the sets at each iteration.

This iterative expansion differs from classical $k$-NN-based PU approaches,
which typically only propagate positive labels.
Here, both classes are progressively constructed,
leading to a symmetric refinement of the decision boundary.

\paragraph{Geometric Interpretation.}
Under the cluster assumption,
points belonging to the same class lie on a low-dimensional manifold.
As iterations proceed,
the negative set expands along directions of maximal discrepancy
from the positive support.
This progressively increases the effective separation
between the empirical supports of $P^+$ and $P^-$.

\subsection{Spectral Entropy Stopping Criterion}

A key challenge in iterative pseudo-labeling
is preventing semantic drift.
To detect when the discovered negative set
has saturated its intrinsic structure,
we monitor its spectral complexity.

Let $\Sigma_t$ denote the empirical covariance matrix
of $\phi(x)$ for $x\in\hat{\mathcal{X}}_{U,N}^{(t)}$:
\begin{equation}
\Sigma_t
=
\frac{1}{|\hat{\mathcal{X}}_{U,N}^{(t)}|}
\sum_{x\in\hat{\mathcal{X}}_{U,N}^{(t)}}
(\phi(x)-\mu_t)(\phi(x)-\mu_t)^\top,
\end{equation}
where $\mu_t$ is the mean of the features $\phi(x)$ for $x\in\hat{\mathcal{X}}_{U,N}^{(t)}$.
Let $\{\lambda_1^{(t)},\dots,\lambda_d^{(t)}\}$
be its eigenvalues normalized such that
$\sum_i \lambda_i^{(t)}=1$.

We define the \emph{spectral entropy}:
\begin{equation}
H(\Lambda_t)
=
-\sum_{i=1}^d
\lambda_i^{(t)} \log \lambda_i^{(t)}.
\end{equation}

\paragraph{Interpretation.}
Spectral entropy quantifies the dispersion
of variance across principal directions.
Low entropy indicates concentration on a narrow manifold;
high entropy reflects a richer, more isotropic structure.
As the negative manifold is progressively uncovered,
$H(\Lambda_t)$ increases.

We stop when \(
H(\Lambda_t) < H(\Lambda_{t-1}),
\) indicating that the geometric expansion has stabilized.
This prevents the algorithm from invading
regions where $P^+$ and $P^-$ overlap,
which would artificially decrease
$\|P^+ - P^-\|_{TV}$.

\paragraph{Theoretical Justification.}
This stopping criterion prevents semantic drift. %
As formalized in Lemma~\ref{lemma:spectral_entropy_general}, this inevitably disperses the mixture's spectrum and strictly decreases spectral entropy.

\begin{lemma}[Spectral Entropy Decrease]
\label{lemma:spectral_entropy_general}
Let $\Sigma_N, \Sigma_P \succ 0$ be arbitrary within-class covariance matrices, and let $\mu_P$ and $\mu_N$ be arbitrary means real vectors. Let us assume that $\Delta\mu = \mu_P - \mu_N \neq 0$. For a contaminated mixture $X \sim (1-\varepsilon)\,\mathcal N(\mu_N,\Sigma_N) + \varepsilon\,\mathcal N(\mu_P,\Sigma_P)$, its spectral entropy satisfies $S(\varepsilon) < S(0)$ for all sufficiently small contamination levels $\varepsilon > 0$. 
\end{lemma}
The full proof is provided in  Appendix \ref{appendix:theoretical_justification}.

\subsection{Knowledge Distillation into Deep Networks}

After convergence,
we obtain the pseudo-labeled dataset
$
\mathcal{D}_{anno}
=
\mathcal{X}_P^{(t)}
\cup
\hat{\mathcal{X}}_{U,N}^{(t)}.
$

We train a parametric classifier
$g_\omega$ on top of embeddings $\phi(x)$,
e.g., a multi-layer perceptron.
The training objective is:
\begin{equation}
\mathcal{L}(\omega)
=
-\frac{1}{|\mathcal{D}_{anno}|}
\sum_{(x,\hat y)\in\mathcal{D}_{anno}}
\mathrm{CE}\big(g_\omega(\phi(x)),\hat y\big).
\end{equation}

\paragraph{Why Distillation Helps.}

The $k$-NN procedure is a non-parametric estimator with low bias but high variance.
Distilling its pseudo-labels into a DNN
induces a smoother decision boundary, reducing variance while preserving local geometry. Importantly, the learned DNN is optimized to keep the local geometry that comes from the $k$-NN, allowing for a good separation between positive and negative.

\section{Experimental results}\label{sec:experiments}

In this section, we evaluate our proposed method \textbf{S-PUNA} and test its robustness under severe covariate shift across diverse datasets. We benchmark \textbf{S-PUNA} against classical positive-unlabeled (PU) risk minimization estimators as well as neighborhood-based baseline methods. Complete implementation details and hyperparameters are provided in the Appendix \ref{appendix:Hyperparameters}. In Section \ref{sec:sup_unsup}, we analyze 21 methods using ResNet and ViT architectures to evaluate how supervised and unsupervised techniques perform in detecting covariate shift. We provide some examples of downstream applications of  \textbf{S-PUNA} in Appendix \ref{appendix:Downstream}.

\subsection{Datasets and Metrics}

We evaluate our proposed method across three primary image classification benchmarks characterized by significant covariate shifts: \textbf{ImageNet-1K} \cite{deng2009imagenet}, \textbf{TinyImageNet} \cite{deng2009imagenet}, and \textbf{EuroSAT} \cite{helber2019eurosat}.

\textit{ImageNet-1K Benchmarks.} Using ImageNet-1K as the Positive Domain (PD) baseline, we test our approach on four shifted datasets. \textbf{ImageNet-V2} \cite{recht2019imagenet} (near shift) replicates the original data collection process while introducing subtle statistical biases. \textbf{ImageNet-R} \cite{hendrycks2021many} (far shift) provides adversarial and rendered versions (e.g., paintings, sketches) of ImageNet classes, preserving semantic content while altering the visual domain. \textbf{ImageNet-C} (far shift)\cite{hendrycksbenchmarking} applies 15 types of algorithmic perturbations, including noise, blur, and weather effects, across five severity levels. Finally, \textbf{GenImage} (near shift) \cite{zhu2023genimage} consists of ImageNet classes regenerated via various generative models, introducing synthetic distributional shifts.

\textit{TinyImageNet Benchmarks.} For TinyImageNet (PD), we consider two shifted variants: \textbf{TinyImageNet-C} \cite{hendrycksbenchmarking} (far shift), which applies corruptions similar to ImageNet-C, and \textbf{TinyImageNet-V2} (near shift), a filtered version of ImageNet-V2 containing only the 200 overlapping classes.

\textit{EuroSAT Benchmarks.} To demonstrate the versatility of our method beyond object-centric domains, we evaluate it on the EuroSAT satellite dataset (PD). We consider two shifted variants: \textbf{EuroSAT-C}(far shift) \cite{oehri2024genformer}, which simulates synthetic atmospheric and sensor-based degradations, and a new dataset introduced in this study, \textbf{EuroSAT-D}(near shift). The latter consists of regenerated versions of EuroSAT images to address synthetic generative shifts. Further details on dataset construction are provided in Appendix \ref{appendix:eurosat} .

\textit{Metrics and Evaluation.} We report standard covariate shift detection metrics: Area Under the Receiver Operating Characteristic curve (\textbf{AUROC}), Area Under the Precision-Recall curve (\textbf{AUPR}), and the False Positive Rate at a 95\% True Positive Rate (\textbf{FPR95}). 

In our analysis, we distinguish between \textbf{near-shift} datasets, which exhibit moderate covariate shifts that are distributionally close to the PD data (and thus more challenging to detect), and \textbf{far-shift} datasets, which represent severe shifts that are more easily identifiable. We provide a comprehensive discussion on this taxonomy in Appendix \ref{appendix:near_vs_far_covariate}.

\subsection{Baselines and Implementation Details}

For evaluation, images are projected into a dense feature space extracted from the sixth block of a ViT pre-trained on in-domain data. We select this specific intermediate layer as it captures structural nuances of covariate shifts while mitigating the task-specific overfitting prevalent in deeper layers (see Appendix \ref{appendix:vit_layer_selection_covariate_shifts}). 

We benchmark \textbf{S-PUNA} against several state-of-the-art PU learning methods, including \textbf{Dist-PU} \cite{zhao2022distpu}, \textbf{DC-PU} \cite{Li2025DCPU}, and \textbf{saPU} \cite{Dai2025CloserLook}, all of which are optimized using the same ViT feature embeddings. Additionally, we compare our approach against pseudo-labeling strategies such as \textbf{LaGAM} \cite{long2024lagam} and a \textbf{naive $k$-NN baseline} \cite{zhang2009reliable} that propagates labels from the positive set. 

For S-PUNA, the iterative expansion process terminates when the absolute difference in spectral entropy between consecutive steps, $H(\Lambda_{t}) - H(\Lambda_{t-1})$, falls below the threshold of $0.0$. An ablation study regarding all of the hyperparameters of  \textbf{S-PUNA} is provided in Appendix \ref{appendix:ablation_study_convergence}. Upon convergence, the learned structural information is distilled into an MLP via standard cross-entropy training. Results using a ResNet instead of the ViT are detailed in Appendix \ref{appendix:resnet}.

\subsection{From supervised to unsupervised Covariate shift detection }
\label{sec:sup_unsup}

To assess the effectiveness of various detection paradigms, we conduct an extensive benchmark in Appendix \ref{appendix:ood_detection_covariateshift}, evaluating 14 OOD detection methods, 5 anomaly detection methods, and 3 diffusion-based techniques across both ResNet and ViT architectures. We further compare these against two supervised approaches: \textit{DiscoPatch} \cite{caetano2025discopatch} and \textit{DRCT} \cite{chen2024drct}. 
\begin{wrapfigure}[12]{r}{0.45\textwidth}
    \centering
     \vspace{-25pt}
        \includegraphics[width=0.5\textwidth]{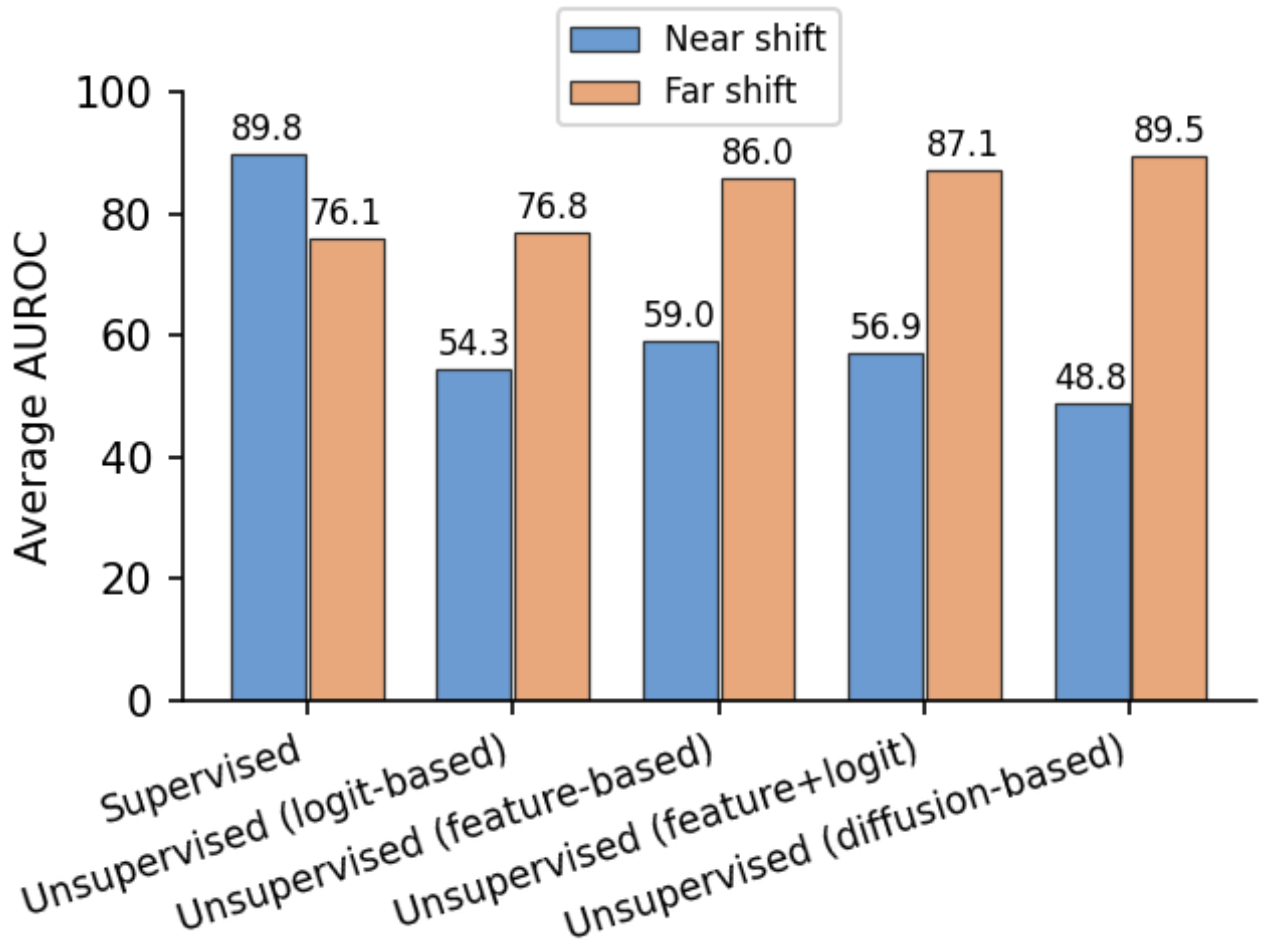}
    \caption{Overview of the performance of ood detection methods on covariate shifted datasets of imagenet1k}
    \label{fig:ood_methods}
\end{wrapfigure}
Our results indicate that supervised techniques achieve superior performance on \textbf{near-shift} scenarios, with an average AUROC of 89.8\%. However, on \textbf{far-shift} datasets, these methods are outperformed by the majority of unsupervised techniques. Among the unsupervised approaches, logit-based methods such as \textit{ReAct} \cite{sun2021react} and \textit{ASH} \cite{djurisicextremely} yield the weakest results, suggesting that high-level class probabilities often discard the subtle structural cues necessary for robust detection. While feature-based methods like \textit{MDS} \cite{lee2018simple}, hybrid approaches like \textit{ViM} \cite{wang2022vim} and diffusion based methods like DiffPath \cite{heng2024out} show marginal improvements, these findings highlight that current OOD detection methods struggle on covariate shift when restricted to purely unsupervised internal signals.

\subsection{Results}

\textit{ImageNet.} 
Table \ref{tab:imagenet} summarizes the covariate shift detection performance on ImageNet-1K for both near and far shift scenarios. 

Classical PU-based baselines exhibit a significant performance gap between the two regimes. For instance, \textbf{Dist-PU} improves from $84.48\%$ AUROC in near-shift cases to $96.70\%$ in far-shift cases, confirming that the latter is considerably easier to identify. However, its near-shift performance remains insufficient, with a high FPR95 of $35.05\%$. Similar trends are observed for \textbf{saPU} and \textbf{LaGAM}, while \textbf{DC-PU} struggles across both regimes, particularly regarding FPR95 ($85.80\%$ near, $89.38\%$ far).

In contrast, \textbf{S-PUNA} consistently outperforms all baselines. Under near-shift conditions, our method achieves $97.89\%$ AUROC—surpassing the strongest baseline (\textbf{Dist-PU}) by over $13\%$—while simultaneously reducing the FPR95 from $35.05\%$ to $9.69\%$ (a $-25.3\%$ improvement). \textbf{S-PUNA} also achieves the highest AUROC under far-shift ($98.51\%$). On average, our approach yields $98.20\%$ AUROC (a $+7.61\%$ gain over \textbf{Dist-PU}) and reduces the average FPR95 to $8.52\%$, outperforming the best baseline by nearly $18\%$. Unlike prior methods, our approach maintains high stability across both shift types, demonstrating superior robustness.
\begin{table}[!htbp]
\centering
\caption{Comparison of methods on \textbf{Imagenet} with \textbf{Near Covariate Shift} and \textbf{Far Covariate Shift} (\textbf{w/ C}: with classifier; \textbf{w/o C}: without classifier; \textbf{C}: classifier). We report AUROC, AUPR, and FPR95 (\%). Results are shown as \textit{mean} $\pm$ \textit{std}.}
\label{tab:imagenet}
\resizebox{\textwidth}{!}{
\begin{tabular}{l ccc ccc ccc}
\toprule
\multirow{2}{*}{\textbf{Method}} & \multicolumn{3}{c}{\textbf{Near Shift}} & \multicolumn{3}{c}{\textbf{Far Shift}} & \multicolumn{3}{c}{\textbf{Average}} \\
\cmidrule(lr){2-4} \cmidrule(lr){5-7} \cmidrule(lr){8-10}
 & AUROC$\uparrow$ & AUPR$\uparrow$ & FPR95$\downarrow$ & AUROC$\uparrow$ & AUPR$\uparrow$ & FPR95$\downarrow$ & AUROC$\uparrow$ & AUPR$\uparrow$ & FPR95$\downarrow$ \\
\midrule
$k$-NN \cite{zhang2009reliable}     & 54.40$\pm$0.00&90.92$\pm$0.00&62.63$\pm$0.00&74.79$\pm$0.00&60.78$\pm$0.00&80.84$\pm$0.00&64.59$\pm$0.00&75.85$\pm$0.00&71.73$\pm$0.00 \\
Dist-PU \cite{zhao2022distpu}       & 84.48$\pm$0.40 & 87.33$\pm$0.30 & 35.05$\pm$0.93 & \underline{96.70$\pm$0.11} & 96.51$\pm$0.14 & 18.06$\pm$0.58 & 90.59$\pm$0.25 & 91.92$\pm$0.22 & 26.56$\pm$0.75 \\
saPU \cite{Dai2025CloserLook} & 82.93$\pm$0.27 & 85.83$\pm$0.26 & 69.58$\pm$1.01 & 95.09$\pm$0.46 & 87.53$\pm$3.17 & 26.78$\pm$2.33 & 89.01$\pm$0.37 & 86.68$\pm$1.72 & 48.18$\pm$1.67 \\
Dc-PU \cite{Li2025DCPU} & 60.61$\pm$2.44 & 65.26$\pm$1.70 & 85.80$\pm$1.03 & 70.51$\pm$3.96 & 41.98$\pm$4.34 & 89.38$\pm$10.83 & 65.56$\pm$3.20 & 53.62$\pm$3.02 & 87.59$\pm$5.93 \\
LaGAM \cite{long2024lagam} & 83.01$\pm$0.54 & 86.40$\pm$0.56 & 65.30$\pm$1.96 & 96.50$\pm$0.13 & 94.36$\pm$1.66 & \underline{17.46$\pm$0.71} & 89.75$\pm$0.33 & 90.38$\pm$1.11 & 41.38$\pm$1.33 \\

\midrule
\rowcolor{lightgreen}\textbf{S-PUNA} w/o C (ours) & \underline{95.81$\pm$0.00}&\underline{97.66$\pm$0.00}&\underline{19.22$\pm$0.00}&95.95$\pm$0.00&\underline{97.93$\pm$0.00}&19.92$\pm$0.00&\underline{95.88$\pm$0.00}&\underline{97.79$\pm$0.00}&\underline{19.57$\pm$0.00} \\
\rowcolor{lightgreen}\textbf{S-PUNA} w/ C (ours) & \textbf{97.89$\pm$0.13}&\textbf{98.81$\pm$0.08}&\textbf{9.69$\pm$0.48}&\textbf{98.51$\pm$0.08}&\textbf{99.20$\pm$0.05}&\textbf{7.36$\pm$0.49}&\textbf{98.20$\pm$0.10}&\textbf{99.00$\pm$0.06}&\textbf{8.52$\pm$0.48} \\
\bottomrule
\end{tabular}
}    
\end{table}
\newline\textit{TinyImageNet.}  
On the TinyImageNet benchmark (Table~\ref{tab:tinyimagenet}), most baselines perform exceptionally well under near-shift conditions; for instance, \textbf{Dist-PU} achieves $99.96\%$ AUROC and $0.15\%$ FPR95. \textbf{S-PUNA} surpasses these results, reaching perfect near-shift performance ($100.00\%$ AUROC, $0.00\%$ FPR95). Our method also provides the strongest far-shift results ($99.64\%$ AUROC, $1.20\%$ FPR95), slightly improving upon the best-performing baseline, \textbf{LaGAM} ($99.59\%$ AUROC, $1.39\%$ FPR95). On average, \textbf{S-PUNA} consistently improves all metrics across the board compared to the state-of-the-art.
\begin{table}[!h]
\centering
\caption{Comparison of methods on \textbf{Tiny-ImageNet} with \textbf{Near Covariate Shift} and \textbf{Far Covariate Shift} (\textbf{w/ C}: with classifier; \textbf{w/o C}: without classifier; \textbf{C}: classifier). We report AUROC, AUPR, and FPR95 (\%). Results are shown as \textit{mean} $\pm$ \textit{std}.}
\label{tab:tinyimagenet}
\resizebox{0.9\textwidth}{!}{
\begin{tabular}{l ccc ccc ccc}
\toprule
\multirow{2}{*}{\textbf{Method}} & \multicolumn{3}{c}{\textbf{Near Shift}} & \multicolumn{3}{c}{\textbf{Far Shift}} & \multicolumn{3}{c}{\textbf{Average}} \\
\cmidrule(lr){2-4} \cmidrule(lr){5-7} \cmidrule(lr){8-10}
 & AUROC$\uparrow$ & AUPR$\uparrow$ & FPR95$\downarrow$ & AUROC$\uparrow$ & AUPR$\uparrow$ & FPR95$\downarrow$ & AUROC$\uparrow$ & AUPR$\uparrow$ & FPR95$\downarrow$ \\
\midrule
$k$-NN \cite{zhang2009reliable}     & 82.00$\pm$0.00&50.65$\pm$0.00&78.39$\pm$0.00&95.82$\pm$0.00&28.19$\pm$0.00&96.96$\pm$0.00&88.91$\pm$0.00&39.42$\pm$0.00&87.68$\pm$0.00 \\
Dist-PU \cite{zhao2022distpu}       & \underline{99.96$\pm$0.00} & \underline{99.98$\pm$0.00} & 0.15$\pm$0.04 & 99.46$\pm$0.11 & 96.14$\pm$0.62 & 2.81$\pm$0.60 & 99.71$\pm$0.06 & 98.06$\pm$0.31 & 1.48$\pm$0.32 \\
saPU \cite{Dai2025CloserLook} & 91.73$\pm$0.01 & 96.37$\pm$0.01 & 19.14$\pm$0.01 & 92.07$\pm$0.17 & 71.47$\pm$1.97 & 33.44$\pm$0.40 & 91.90$\pm$0.09 & 83.92$\pm$0.99 & 26.29$\pm$0.21\\
Dc-PU \cite{Li2025DCPU} & 95.93$\pm$2.90 & 97.82$\pm$1.52 & 16.43$\pm$13.11 & 82.95$\pm$5.58 & 35.21$\pm$14.85 & 61.76$\pm$21.21 & 89.44$\pm$4.24 & 66.52$\pm$8.19 & 39.09$\pm$17.16 \\
LaGAM \cite{long2024lagam}     & 99.93$\pm$0.01 & 99.96$\pm$0.01 & 0.22$\pm$0.01 & \underline{99.59$\pm$0.21} & 95.35$\pm$3.03 & \underline{1.39$\pm$0.78} & \underline{99.76$\pm$0.11} & 97.66$\pm$1.52 & 0.81$\pm$0.39 \\

\midrule
\rowcolor{lightgreen}\textbf{S-PUNA} w/o C (ours) & \textbf{100.00$\pm$0.00}&\textbf{100.00$\pm$0.00}&\textbf{0.00$\pm$0.00}&99.54$\pm$0.00&\textbf{99.62$\pm$0.00} & \textbf{1.20$\pm$0.00} & \textbf{99.77$\pm$0.00}&\textbf{99.81$\pm$0.00}&\textbf{0.60$\pm$0.00} \\
\rowcolor{lightgreen}\textbf{S-PUNA} w/ C (ours) & 99.90$\pm$0.10&99.92$\pm$0.08&\underline{0.02$\pm$0.04} & \textbf{99.64$\pm$0.11} & \underline{99.46$\pm$0.15}&1.51$\pm$0.50&\textbf{99.77$\pm$0.11}&\underline{99.69$\pm$0.12}&\underline{0.77$\pm$0.27} \\

\bottomrule
\end{tabular}
} 
\end{table}
\newline\textit{EuroSAT.}  
Results on the EuroSAT dataset (Table~\ref{tab:eurosat}) reveal a critical vulnerability in traditional PU methods. While several baselines achieve high AUROC under near-shift conditions (\textbf{Dist-PU}: $99.78\%$), their performance degrades substantially under far-shift scenarios—\textbf{Dist-PU}, for example, drops to $89.46\%$ AUROC with a significant $49.97\%$ FPR95. In contrast, \textbf{S-PUNA} achieves perfect separation in the far-shift regime ($100.00\%$ AUROC, $0.00\%$ FPR95), marking a $+10.54\%$ AUROC gain and a $-49.97\%$ reduction in FPR95 over the best baseline. Averaged across both regimes, \textbf{S-PUNA} reaches $99.90\%$ AUROC and $0.33\%$ FPR95, substantially outperforming \textbf{LaGAM}'s average metrics ($92.79\%$ AUROC, $28.92\%$ FPR95). These results confirm the superior robustness of our geometry-aware approach across varying shift severities. Please refer to Appendix \ref{appendix:gen} for transfer experiments and robustness across shifts.

\begin{table}[h]
\centering
\vspace{-10pt}
\caption{Comparison of methods on \textbf{EuroSAT} with \textbf{Near} and \textbf{Far Covariate Shift}. (\textbf{w/ C}: with classifier; \textbf{w/o C}: without classifier; \textbf{C}: classifier). We report AUROC, AUPR, and FPR95 (\%). Results are shown as \textit{mean} $\pm$ \textit{std}.}
\label{tab:eurosat}
\resizebox{0.9\textwidth}{!}{
\begin{tabular}{l ccc ccc ccc}
\toprule
\multirow{2}{*}{\textbf{Method}} & \multicolumn{3}{c}{\textbf{Near Shift}} & \multicolumn{3}{c}{\textbf{Far Shift}} & \multicolumn{3}{c}{\textbf{Average}} \\
\cmidrule(lr){2-4} \cmidrule(lr){5-7} \cmidrule(lr){8-10}
 & AUROC$\uparrow$ & AUPR$\uparrow$ & FPR95$\downarrow$ & AUROC$\uparrow$ & AUPR$\uparrow$ & FPR95$\downarrow$ & AUROC$\uparrow$ & AUPR$\uparrow$ & FPR95$\downarrow$ \\
\midrule
$k$-NN \cite{zhang2009reliable}     & 58.86$\pm$0.00&84.80$\pm$0.00&13.91$\pm$0.00&81.03$\pm$0.00&83.11$\pm$0.00&58.47$\pm$0.00&69.95$\pm$0.00&83.96$\pm$0.00&36.19$\pm$0.00 \\
Dist-PU \cite{zhao2022distpu}       & \underline{99.78$\pm$0.13} & \textbf{99.95$\pm$0.03} & \underline{0.72$\pm$0.46} & \underline{89.46$\pm$0.50} & \underline{96.48$\pm$0.26} & \underline{49.97$\pm$2.81} & 94.62$\pm$0.31 & \textbf{98.22$\pm$0.14} & \underline{25.35$\pm$1.64}\\
saPU \cite{Dai2025CloserLook} & 96.71$\pm$0.93 & 99.30$\pm$0.21 & 7.55$\pm$3.58 & 86.75$\pm$0.62 & 95.81$\pm$0.15 & 54.38$\pm$1.06 & 91.73$\pm$0.78 & 97.56$\pm$0.18 & 30.97$\pm$2.32 \\
Dc-PU \cite{Li2025DCPU} & 68.96$\pm$3.53 & 87.87$\pm$0.59 & 90.89$\pm$6.73 & 60.22$\pm$4.34 & 85.05$\pm$1.43 & 94.48$\pm$3.06 & 64.59$\pm$3.93 & 86.46$\pm$1.01 & 92.68$\pm$4.90 \\
LaGAM \cite{long2024lagam} & 99.33$\pm$0.56 & \underline{99.86$\pm$0.14} & 3.01$\pm$3.39 & 86.24$\pm$0.90 & 96.28$\pm$0.27 & 54.83$\pm$3.24 & \underline{92.79$\pm$0.73} & \underline{98.07$\pm$0.21} & 28.92$\pm$3.32\\

\midrule
\rowcolor{lightgreen}\textbf{S-PUNA} w/o C (ours) & 84.40$\pm$0.00&38.02$\pm$0.00&55.76$\pm$0.00&\textbf{100.00$\pm$0.00}&\textbf{100.00$\pm$0.00}&\textbf{0.00$\pm$0.00}&92.20$\pm$0.00&69.01$\pm$0.00&27.88$\pm$0.00 \\
\rowcolor{lightgreen}\textbf{S-PUNA} w/ C (ours) & \textbf{99.79$\pm$0.05} & 96.08$\pm$0.70&\textbf{0.65$\pm$0.36}&\textbf{100.00$\pm$0.00}&\textbf{100.00$\pm$0.00}&\textbf{0.00$\pm$0.00}&\textbf{99.90$\pm$0.03}&98.04$\pm$0.35&\textbf{0.33$\pm$0.18} \\

\bottomrule
\end{tabular}
} 
\end{table}
\vspace{-3mm}
\subsection{Ablation and Sensitivity Analysis}
\textit{Classifier head contribution.} Across all datasets, we observe that the classifier head often improves over the S-PUNA w/o C variant; when it does not, its performance remains comparable within the metric $\pm$ the standard deviation.
\newline
\textit{Effect of the proportion of shifted data.}
Previous experiments considered the standard 1:1 composition of the unlabeled pool, corresponding to \textbf{50\%} shifted samples in Table~\ref{tab:prevalence_auroc}. Using the same features extracted from block 6 of the ViT, we conducted additional experiments on ImageNet-1K and its associated near- and far-shifted datasets with shifted-sample proportions of \textbf{5\%}, \textbf{25\%}, and \textbf{75\%}. Table~\ref{tab:prevalence_auroc} reports the average AUROC for covariate-shift detection across all datasets. S-PUNA consistently outperforms the PU-learning baselines for every considered proportion of shifted data.
\newline
\textit{Effect of ViT layer selection.}
We also compared features extracted from blocks 6 and 11 of the ViT for shift detection on ImageNet-1K, while keeping the S-PUNA pipeline and all hyperparameters unchanged. The corresponding results are reported in the first column of Table~\ref{tab:prevalence_auroc}. S-PUNA continues to outperform the PU-learning baselines when block-11 features are used, indicating that its improvement is not merely an artifact of selecting block 6. Nevertheless, block 6 provides the best overall performance. Additional results are in Appendix \ref{appendix:vit_layer_selection_covariate_shifts}.

\begin{table}[t]
\centering
\scriptsize
\setlength{\tabcolsep}{3pt}
\renewcommand{\arraystretch}{0.82}
\caption{AUROC across different shifted-sample ratios in INet1K.}
\label{tab:prevalence_auroc}
\begin{tabular}{lccccc}
\toprule
\multirow{3}{*}{Method} & \multicolumn{5}{c}{Average AUROC} \\
\cmidrule(lr){2-6}
 & \multirow{2}{*}{Block 11} & \multicolumn{4}{c}{Block 6} \\
\cmidrule(lr){3-6}
 &  & $p=5\%$ & $p=25\%$ & $p=50\%$ & $p=75\%$ \\
\midrule
DCPU       & 62.23 & 73.08 & 68.00 & 65.56 & 72.19 \\
DistPU     & 82.08 & 71.04 & 86.40 & 90.59 & 92.17 \\
LaGAM      & 79.35 & 85.08 & 87.45 & 89.75 & 90.92 \\
SAPU       & 79.38 & 60.24 & 75.53 & 89.01 & 92.03 \\
S-PUNA w/c & \textbf{97.03} & \textbf{91.61} & \textbf{95.75} & \textbf{98.20} & \textbf{95.53} \\
\bottomrule
\end{tabular}
\vspace{-5mm}
\end{table}

\section{Conclusion}

In this work, we investigated the efficacy of various paradigms for covariate shift detection. Our analysis demonstrates that while fully supervised approaches perform reasonably well, traditional unsupervised covariate shift detection methods—relying on internal signals like logits or feature statistics fail significantly in the near shift regime. Because covariate shifts preserve semantic content and introduce only subtle distributional changes, these unsupervised signals are unable to reliably distinguish shifted samples from in-distribution data.

To bridge this gap, we proposed a weakly supervised alternative based on the Positive–Unlabeled (PU) learning framework. To mitigate the instability of classical PU risk estimators under significant distributional overlap, we introduced \textbf{Spectral PU Neighborhood Annotation (S-PUNA)}. This geometry-aware method progressively discovers shifted samples through local neighborhood expansion within the feature manifold.

Extensive experiments across multiple benchmarks confirm that \textbf{S-PUNA} consistently outperforms existing PU baselines and even surpasses fully supervised models in several settings. Our findings suggest that strictly supervised signals are not a prerequisite for reliable covariate shift detection; rather, by effectively leveraging the intrinsic geometric structure of feature representations, weakly supervised PU learning can achieve highly robust and state-of-the-art performance.

\textbf{Acknowledgments.}
This work was granted access to the HPC resources of IDRIS under the allocation AD011015965R1 and AD011016949 made by GENCI.

\clearpage  %

\clearpage
\appendix
\setcounter{page}{1}
\setcounter{section}{0}
\setcounter{subsection}{0}

\renewcommand{\theHsection}{supp.\thesection}
\renewcommand{\theHsubsection}{supp.\thesubsection}
\renewcommand{\theHtable}{supp.\thetable}
\renewcommand{\theHfigure}{supp.\thefigure}

\renewcommand{\thetable}{A.\arabic{table}}
\renewcommand{\thefigure}{A.\arabic{figure}}

\begin{center}
  \Large \textbf{From Local Geometry to Global Pseudo-Labeling for Robust Positive–Unlabeled Learning under Covariate Shift\\ --Supplementary Material--}
\end{center}

\vspace{0.8cm}

\setcounter{tocdepth}{2}
\providecommand{\authcount}[1]{}
\startcontents[supp]

{
\hypersetup{linkcolor=black}
\printcontents[supp]{}{1}{\section*{\centering Table of Contents}}{}
}

\vspace{0.5cm}

\clearpage

\section{Theoretical Justification of the Spectral Entropy Criterion}\label{appendix:theoretical_justification}

\subsection{Contamination Model for the Annotated Negative Set}\label{appendix:contamination model}

To justify the spectral entropy stopping rule, we propose \textbf{Lemma 1} in the main paper. In this section, we analyze the evolution of the covariance structure of the progressively discovered negative set to prove this Lemma.

Assume that at iteration $t$, the annotated negative set is composed of:
\begin{itemize}
    \item true negatives drawn from $\mathcal N(\mu_N,\Sigma_N)$ with probability $1-\varepsilon$,
    \item contaminated positives drawn from $\mathcal N(\mu_P,\Sigma_P)$ with probability $\varepsilon$,
\end{itemize}
where $\varepsilon \in [0,1]$ denotes the contamination level induced by pseudo-labeling errors.

The resulting population distribution is therefore the Gaussian mixture

\begin{equation}
X \sim (1-\varepsilon)\,\mathcal N(\mu_N,\Sigma_N)
+ \varepsilon\,\mathcal N(\mu_P,\Sigma_P)    
\end{equation}

\paragraph{Population covariance of the contaminated set.}

Let $\Delta\mu := \mu_P - \mu_N$. The covariance of a two-component Gaussian mixture satisfies the classical decomposition:

\begin{equation}
\Sigma(\varepsilon)
=
(1-\varepsilon)\Sigma_N
+
\varepsilon\Sigma_P
+
\varepsilon(1-\varepsilon)\Delta\mu\Delta\mu^\top.
\end{equation}

The first two terms correspond to the within-class covariance, while the third term is the \emph{between-class covariance}, which is positive semi-definite and of rank at most $1$.

Define the normalized eigenvalues of $\Sigma(\varepsilon)$:
\begin{equation}
\rho_i(\varepsilon)
=
\frac{\lambda_i(\Sigma(\varepsilon))}
{\operatorname{tr}(\Sigma(\varepsilon))},
\qquad
\sum_{i=1}^d \rho_i(\varepsilon)=1.
\end{equation}

The spectral entropy is defined as

\begin{equation}
S(\varepsilon)
=
-\sum_{i=1}^d
\rho_i(\varepsilon)\log \rho_i(\varepsilon).
\end{equation}

We now show that $S(\varepsilon)$ strictly decreases as soon as $\varepsilon>0$, which formally justifies entropy-based stopping.

\subsection{Special Case: Isotropic Covariances}

Assume
\begin{equation}
\Sigma_N=\Sigma_P=\sigma^2 I_d.
\end{equation}

Then

\begin{equation}
\Sigma(\varepsilon)
=
\sigma^2 I_d
+
\varepsilon(1-\varepsilon)\Delta\mu\Delta\mu^\top.
\end{equation}

Let
\begin{equation}
\alpha(\varepsilon)
=
\varepsilon(1-\varepsilon)\|\Delta\mu\|^2,
\qquad
u=\frac{\Delta\mu}{\|\Delta\mu\|}.
\end{equation}

Then
\begin{equation}
\Sigma(\varepsilon)
=
\sigma^2 I_d
+
\alpha(\varepsilon)uu^\top,
\end{equation}
which is a rank-1 perturbation of a scaled identity.

\paragraph{Eigenvalues.}

By the spectral theorem for rank-1 perturbations,

\begin{equation}
\lambda_1(\varepsilon)
=
\sigma^2+\alpha(\varepsilon),
\qquad
\lambda_i(\varepsilon)=\sigma^2 \quad (i=2,\dots,d).
\end{equation}

The trace equals

\begin{equation}
\operatorname{tr}(\Sigma(\varepsilon))
=
d\sigma^2+\alpha(\varepsilon).
\end{equation}

The normalized eigenvalues are therefore

\begin{equation}
\rho_1(\varepsilon)
=
\frac{\sigma^2+\alpha(\varepsilon)}
{d\sigma^2+\alpha(\varepsilon)},
\qquad
\rho_i(\varepsilon)
=
\frac{\sigma^2}
{d\sigma^2+\alpha(\varepsilon)}.
\end{equation}

\paragraph{Pure case.}

If $\varepsilon=0$, then $\alpha(0)=0$ and

\begin{equation}
\rho_i(0)=\frac{1}{d}
\quad \forall i,
\end{equation}

which yields

\begin{equation}
S(0)
=
-\sum_{i=1}^d \frac{1}{d}\log\frac{1}{d}
=
\log d,
\end{equation}

the maximal entropy over the probability simplex.

\paragraph{Contaminated case.}

For any $\varepsilon\in(0,1)$ with $\|\Delta\mu\|>0$, we have $\alpha(\varepsilon)>0$, implying

\begin{equation}
\rho_1(\varepsilon)>\frac{1}{d}
>
\rho_i(\varepsilon) \quad (i\ge2).
\end{equation}

Hence $\rho(\varepsilon)$ is non-uniform and strictly majorizes the uniform vector.

Since Shannon entropy is strictly Schur-concave (because $-x\log x$ is strictly concave), strict majorization implies

\begin{equation}
S(\varepsilon)<S(0).
\end{equation}

Moreover, $S(\varepsilon)$ is continuous in $\varepsilon$, and a first-order Taylor expansion around $0$ shows

\begin{equation}
S(\varepsilon)
=
\log d
-
C\varepsilon
+
o(\varepsilon),
\quad C>0,
\end{equation}

so entropy decreases immediately at infinitesimal contamination.

\paragraph{Interpretation.}

Even arbitrarily small contamination injects a directional variance component, concentrating spectral mass on the leading eigenvalue and reducing entropy. Thus entropy is maximized at purity and strictly decreases once semantic drift begins.

\subsection{General Case: Arbitrary Covariances}

Let
\begin{equation}
A(\varepsilon)
=
(1-\varepsilon)\Sigma_N
+
\varepsilon\Sigma_P,    
\end{equation}

\begin{equation}
B(\varepsilon)
=
\varepsilon(1-\varepsilon)\Delta\mu\Delta\mu^\top
\succeq 0.
\end{equation}

Then

\begin{equation}
\Sigma(\varepsilon)
=
A(\varepsilon)+B(\varepsilon).
\end{equation}

\paragraph{Eigenvalue monotonicity.}

By Weyl’s monotonicity theorem for Hermitian matrices,

\begin{equation}
\lambda_i(\Sigma(\varepsilon))
\ge
\lambda_i(A(\varepsilon))
\quad \forall i.
\end{equation}

Since $\operatorname{tr}(B(\varepsilon))>0$ for $\varepsilon\in(0,1)$, the largest eigenvalue satisfies

\begin{equation}
\lambda_1(\Sigma(\varepsilon))
>
\lambda_1(A(\varepsilon)).
\end{equation}

Thus the spectrum of $\Sigma(\varepsilon)$ becomes strictly more dispersed.

\paragraph{Majorization argument.}

Let $\tilde{\rho}(\varepsilon)$ be the normalized eigenvalues of $A(\varepsilon)$ and $\rho(\varepsilon)$ those of $\Sigma(\varepsilon)$. Ky Fan inequalities imply that the partial sums of the top $k$ eigenvalues strictly increase under addition of $B(\varepsilon)$.

Therefore $\rho(\varepsilon)$ strictly majorizes $\tilde{\rho}(\varepsilon)$, and by strict Schur-concavity of Shannon entropy,

\begin{equation}
S(\Sigma(\varepsilon))
<
S(A(\varepsilon)).
\end{equation}

Since $A(\varepsilon)\to\Sigma_N$ as $\varepsilon\to0^+$, we obtain

\begin{equation}
S(\varepsilon)
<
S(0)
\quad \text{for all sufficiently small } \varepsilon>0.
\end{equation}

\subsection{Finite-Sample Concentration}

In practice, we observe the empirical covariance $\hat{\Sigma}_t$. By matrix concentration inequalities (e.g., matrix Bernstein bounds), its eigenvalues converge uniformly to those of $\Sigma(\varepsilon)$:

\begin{equation}
\|\hat{\Sigma}_t-\Sigma(\varepsilon)\|
=
O_p\!\left(\sqrt{\frac{d}{n}}\right).
\end{equation}

Hence the empirical spectral entropy $\hat S_t$ concentrates around $S(\varepsilon)$, ensuring that entropy drop is detectable with high probability before contamination becomes large.

In both isotropic and general Gaussian settings, the mechanism is identical: the Shannon entropy strictly decreases when the N data are contaminated.

Therefore, spectral entropy is maximized when the negative set is pure and strictly decreases as soon as contamination begins. 

This provides a rigorous geometric justification for using spectral entropy as a principled stopping criterion: it detects the earliest onset of semantic drift through spectral anisotropy in feature space.

\section{Details on Baseline Implementations}\label{appendix:baseline_implementation}

All baselines are evaluated under the same detection protocol for covariate shift across three in-distribution datasets: ImageNet, Tiny-ImageNet, and EuroSAT.

For each in-distribution dataset, we first sample a class-balanced positive set $P$ from the training split. We then construct an unlabeled set $U$ by combining two sources of data: (i) positive samples taken from the validation split of the in-distribution dataset (ImageNet validation set, Tiny-ImageNet validation set, or EuroSAT validation set), and (ii) samples originating from the considered covariate-shift dataset. The remaining part of the validation split is used as the test set and is never observed during the training procedure.

All PU baselines are trained using a Multi-Layer Perceptron (MLP) composed of a single hidden layer with 512 neurons. The MLP operates on frozen feature embeddings extracted from a pre-trained backbone network.

We consider two types of feature extractors: a Vision Transformer (ViT) and a ResNet-50. In both cases, the backbone networks are pre-trained on the corresponding in-distribution dataset (ImageNet for ImageNet and Tiny-ImageNet experiments, and EuroSAT for EuroSAT experiments).

\paragraph{ViT features.}
For the Vision Transformer, we extract the CLS token representation from block 6 of the transformer encoder. This produces a feature vector of dimension 768, which is used as the input embedding for the MLP classifier.

\paragraph{ResNet-50 features.}
For ResNet-50, we extract intermediate convolutional features from the output of the second residual block group (also called \texttt{layer2}). The resulting feature map is then processed using global average pooling to produce a fixed-dimensional embedding of size 512, which is used as input to the MLP.

\subsection{Metrics}

Each method is evaluated using three different random seeds. 
We report three standard metrics commonly used in distribution shift detection~\cite{yang2022openood}: 
AUROC, AUPR, and FPR95.

Given the test split, we compute the \textbf{AUROC}, which measures the ability of a model to distinguish between positive and negative samples across all possible decision thresholds. 

We also report the \textbf{AUPR}, which evaluates the trade-off between precision and recall across thresholds.

Finally, we compute \textbf{FPR95}, defined as the false positive rate when the true positive rate reaches $95\%$. Concretely, the shift detection threshold is set at the $(1 - 0.95)$ quantile of the shift (covariate or semantic) scores, and the false positive rate is then measured on the in-distribution samples. Lower FPR95 values indicate better detection performance.

\subsection{Construction of $P$ and $U$}

For each experiment, both the Positive distribution (In-Distribution ID) and covariate shift datasets are shuffled according to the current random seed. We then sample examples from the ID and covariate shift datasets to construct the unlabeled set $U$.

Across all experiments, $U$ is constructed such that the class prior is equal to $0.5$. More precisely,

\[
U = U_{\mathrm{ID}} \cup U_{\mathrm{Covariate}}, 
\quad 
\pi = \frac{|U_{\mathrm{ID}}|}{|U_{\mathrm{ID}}| + |U_{\mathrm{Covariate}}|} = 0.5.
\]

For each dataset, the number of samples used for $P$, $U_{\mathrm{ID}}$, and $U_{\mathrm{Covariate}}$ is summarized in Table~\ref{tab:PU_samples_table}.

All remaining samples that are not used to construct $U$ are used as test data for computing AUROC, AUPR, and FPR95.

For corrupted datasets (ImageNet-C, Tiny-ImageNet-C, and EuroSAT-C), we evaluate each corruption severity separately. For a given severity level, we pool all corruption types associated with that severity.

\begin{table}[t]
\centering
\caption{\textbf{Number of samples used for $P$ and $U$ across datasets.}}
\label{tab:PU_samples_table}
\begin{tabular}{lcccc}
\toprule
Dataset & $P$ & $U_{ID}$ & $U_{Covariate}$ & $U$ \\
\midrule
ImageNet & 1000 & 7000 & 7000 & 14000 \\
Tiny-ImageNet & 1000 & 2000 & 2000 & 4000 \\
EuroSAT & 1000 & 1000 & 1000 & 2000 \\
\bottomrule
\end{tabular}
\end{table}

\subsection{Hyperparameters}
\label{appendix:Hyperparameters}

Unless otherwise stated, all PU baselines in the main paper are trained using ViT features extracted from the CLS token of the sixth transformer block of a ViT-B/16 backbone \cite{dosovitskiyimage}. We use the same MLP for all the baselines and the same embedding, and exactly the same data. 

\paragraph{Dist-PU.}
We use 30 warm-up epochs followed by 150 PU training epochs for ImageNet, and 15 warm-up and 15 PU epochs for EuroSAT and Tiny-ImageNet. The training batch size is 2048, and the learning rate is $10^{-4}$. The entropy, Mixup, and mixed-entropy coefficients are set to $\mu=0.01$, $\nu=2.0$, and $\gamma=0.02$, respectively. The Mixup Beta parameter is $\alpha=6.0$.

\paragraph{DC-PU.}
We use the logistic loss and train for 30 epochs on ImageNet and 10 epochs on Tiny-ImageNet and EuroSAT. Each epoch consists of 200 iterations with batch sizes $B_P = B_U = 512$. The learning rate is $3 \times 10^{-3}$ with learning rate milestones at 6000 and 9000 iterations and a decay factor $0.1$. The exponential moving average parameter is $\beta=0.9$. The penalty parameter is $\tau = 10^{-2}$, the rollback margin is $\gamma = 0.7$, and the Lagrange multiplier $\Theta$ is randomly initialized with standard deviation $0.01$.

\paragraph{SAPU.}
Training is performed for 50 epochs with 200 iterations per epoch, using batch sizes $B_P = B_U = 512$ and $n_{\mathrm{bag}} = 4$. We use the logistic loss and the AdamW optimizer with learning rate $3\times10^{-4}$ and weight decay $10^{-4}$.

\paragraph{LaGAM.}
For meta-validation, we use 250 in-distribution samples and 250 shifted samples. Training is performed for 80 epochs on ImageNet and 50 epochs on EuroSAT and Tiny-ImageNet. The batch size is 1024, and the meta-batch size is 256. Optimization is performed with AdamW using a learning rate $3\times10^{-4}$ and weight decay $10^{-4}$. The EMA coefficient for pseudo-label updates is linearly scheduled from $0.95$ to $0.8$. The meta-update is applied to the layer labeled \texttt{classifier}. The auxiliary Mixup loss uses weight $1.0$ and Beta parameter $\alpha=4.0$.

\subsection{EuroSAT and Variant Datasets}
\label{appendix:eurosat}

The EuroSAT dataset~\cite{helber2019eurosat} is a widely used benchmark for land use and land cover classification. It is built from open-access Earth observation data collected by the Copernicus Sentinel-2 satellite. The dataset contains approximately 27,000 labeled and geo-referenced image patches, each covering a $64 \times 64$ pixel area with a spatial resolution of 10 meters per pixel.

In our experiments, we use the RGB version of EuroSAT, which consists of the red, green, and blue spectral bands encoded as standard JPEG images. This format makes the dataset compatible with standard deep learning architectures used in computer vision.

The dataset contains ten classes: \textbf{Annual Crop}, \textbf{Forest}, \textbf{Herbaceous Vegetation}, \textbf{Highway}, \textbf{Industrial Buildings}, \textbf{Pasture}, \textbf{Permanent Crop}, \textbf{Residential Buildings}, \textbf{River}, and \textbf{Sea/Lake}.

\textit{EuroSAT-C.} 
We use EuroSAT-C~\cite{oehri2024genformer} to evaluate robustness under corrupted input conditions. This dataset is derived from the EuroSAT test set by applying a set of synthetic corruptions including algorithmic \textit{noise}, \textit{blur}, \textit{simulated weather effects}, and \textit{digital artifacts}. Each corruption is applied at multiple severity levels.

These perturbations are designed to mimic realistic degradations encountered in satellite imagery, such as atmospheric disturbances or sensor noise. Evaluating models on EuroSAT-C therefore provides a systematic assessment of robustness to domain shifts in remote sensing scenarios.

\textit{EuroSAT-D} 
We also introduce \textbf{EuroSAT-D}, a synthetic dataset generated using a Stable Diffusion model fine-tuned on the EuroSAT RGB training set with Low-Rank Adaptation (LoRA).
This dataset was constructed independently of S-PUNA and was not tuned in any way to favor it over other baselines.

To ensure semantic consistency and visual quality, generated images are filtered using a dual-model validation pipeline. We use a fully supervised ViT-Large model pretrained on ImageNet and fine-tuned on EuroSAT classification. Only generated samples for which both networks predict the target class with confidence greater than $0.7$ are retained.

For the filtering step, the ViT-Large model is fine-tuned on $224 \times 224$ inputs for 20 epochs with a batch size of 64 and a $0.2$ validation split. Training uses a cosine annealing learning rate schedule with a 3-epoch warmup, base learning rate $10^{-4}$, and weight decay $10^{-4}$.

\subsection{Results across Datasets}

Across all datasets, S-PUNA consistently outperforms existing PU-learning baselines and approaches the performance of fully supervised classifiers. 
On the ImageNet benchmark (Table~\ref{tab:Appendiximagenet}), classical PU methods such as Dist-PU, saPU, DC-PU, and LaGAM achieve moderate detection performance, with average AUROC values ranging between $65.56$ and $90.59$. In contrast, S-PUNA significantly improves this performance, reaching an average AUROC of $98.20$ when the classifier is used. This result reduces the gap with the supervised classifier (AUROC $98.61$) to less than $0.5$ points, while operating under the weaker PU supervision setting. A similar trend can be observed for AUPR and FPR95, where S-PUNA drastically reduces the false positive rate compared to existing PU methods.

A key observation is the strong performance of S-PUNA across different types of distribution shifts. In particular, the method remains robust on both near shifts (e.g., ImageNet-V2 or EuroSAT-D) and far shifts (e.g., ImageNet-C or EuroSAT-C). For example, on GenImage and ImageNet-C, S-PUNA achieves AUROC values above $99$, demonstrating that the proposed approach effectively separates shifted samples even under severe covariate shifts. This stability across shifts contrasts with several PU baselines whose performance varies significantly depending on the dataset.

The comparison between the two variants of our method further highlights the benefit of the final classifier distillation step. The variant without classifier (S-PUNA w/o C) already provides strong performance, outperforming most PU baselines on several datasets. However, adding the classifier consistently improves the results, especially on challenging near-shift scenarios. For instance, on ImageNet-V2, the AUROC increases from $92.65$ to $96.38$, and the FPR95 drops from $33.70$ to $17.14$. Similar improvements are observed on EuroSAT-D, where the classifier significantly boosts AUROC from $84.40$ to $99.79$. This confirms that the distillation step helps transform the non-parametric neighborhood structure discovered during the PU phase into a smoother and more robust decision boundary.

Finally, results on Tiny-ImageNet (Table~\ref{tab:tinyimagenet}) and EuroSAT (Table~\ref{tab:eurosat}) confirm the strong generalization of S-PUNA across datasets of different sizes and domains. In particular, S-PUNA often matches or surpasses the best PU baselines while remaining very close to the supervised upper bound. These results suggest that exploiting local neighborhood expansion within the PU framework allows the model to progressively uncover the negative distribution and produce reliable shift detection even with limited labeled supervision.

\begin{table}[t]
\centering
\caption{\textbf{Main Results: Performance comparison across four ImageNet variants and Overall Average}(\textbf{w/ C}: with classifier; \textbf{w/o C}: without classifier; \textbf{C}: classifier). Results are shown as \textit{mean} $\pm$ \textit{std}.}
\label{tab:Appendiximagenet}
\scriptsize
\setlength{\tabcolsep}{2pt} 
\begin{tabular}{l ccc ccc}
\toprule
\multirow{2}{*}{\textbf{Method}} & \multicolumn{3}{c}{\textbf{ImageNet-V2}} & \multicolumn{3}{c}{\textbf{ImageNet-R}} \\
\cmidrule(lr){2-4} \cmidrule(lr){5-7}
 & \scriptsize AUROC\up & \scriptsize AUPR\up & \scriptsize FPR95\down & \scriptsize AUROC\up & \scriptsize AUPR\up & \scriptsize FPR95\down \\
\midrule
KNN \cite{zhang2009reliable} & 51.51$\pm$0.00&93.55$\pm$0.00&66.67$\pm$0.00&62.59$\pm$0.00&88.49$\pm$0.00&73.92$\pm$0.00 \\
Dist-PU \cite{zhao2022distpu}     & 77.60$\pm$0.55&84.69$\pm$0.28&25.08$\pm$0.62&93.59$\pm$0.21&95.90$\pm$0.10&35.48$\pm$1.15 \\
saPU  \cite{Dai2025CloserLook}      & 75.53$\pm$0.22&83.32$\pm$0.43&87.43$\pm$1.63&91.11$\pm$0.64&93.51$\pm$0.59&50.83$\pm$3.94 \\
Dc-PU \cite{Li2025DCPU} & 56.15$\pm$3.85&69.24$\pm$2.32&82.23$\pm$0.70&63.57$\pm$0.85&72.23$\pm$3.86&92.13$\pm$7.92 \\
LaGAM  \cite{long2024lagam}      & 75.77$\pm$1.01&84.43$\pm$0.74&78.49$\pm$1.67&93.50$\pm$0.12&95.95$\pm$0.15&33.11$\pm$1.00 \\
\midrule Supervised classifier & \textbf{97.05$\pm$0.38}&\textbf{98.10$\pm$0.30}&\textbf{13.60$\pm$2.06}&\textbf{97.86$\pm$0.39}&\textbf{98.74$\pm$0.28}&\textbf{9.99$\pm$1.86}\\
\midrule\rowcolor{lightgreen} S-PUNA w/o C (ours) & 92.65$\pm$0.00&95.45$\pm$0.00&33.70$\pm$0.00&93.54$\pm$0.00&96.28$\pm$0.00&30.45$\pm$0.00 \\
\rowcolor{lightgreen} S-PUNA w/ C (ours) & \underline{96.38$\pm$0.20}&\underline{97.69$\pm$0.15}&\underline{17.14$\pm$0.85}&\underline{97.34$\pm$0.11}&\underline{98.49$\pm$0.08}&\underline{13.37$\pm$0.82} \\
\midrule \midrule 

\multirow{2}{*}{\textbf{Method}} & \multicolumn{3}{c}{\textbf{GenImage}} & \multicolumn{3}{c}{\textbf{ImageNet-C}} \\
\cmidrule(lr){2-4} \cmidrule(lr){5-7}
 & \scriptsize AUROC\up & \scriptsize AUPR\up & \scriptsize FPR95\down & \scriptsize AUROC\up & \scriptsize AUPR\up & \scriptsize FPR95\down \\
\midrule
KNN  \cite{zhang2009reliable}  & 57.28$\pm$0.00&88.28$\pm$0.00&58.59$\pm$0.00&86.98$\pm$0.00&33.07$\pm$0.00&87.75$\pm$0.00 \\
Dist-PU \cite{zhao2022distpu}      & 91.36$\pm$0.24&89.96$\pm$0.31&45.02$\pm$1.23&\underline{99.81$\pm$0.00}&97.11$\pm$0.18&\underline{0.65$\pm$0.02} \\
saPU  \cite{Dai2025CloserLook}       & 90.32$\pm$0.32&88.34$\pm$0.10&51.74$\pm$0.39&99.06$\pm$0.29&81.55$\pm$5.74&2.73$\pm$0.72 \\
Dc-PU \cite{Li2025DCPU}  & 65.08$\pm$1.03&61.28$\pm$1.08&89.36$\pm$1.36&77.44$\pm$7.06&11.73$\pm$4.82&86.64$\pm$13.74 \\
LaGAM  \cite{long2024lagam}      & 90.24$\pm$0.06&88.36$\pm$0.38&52.11$\pm$2.24&99.49$\pm$0.13&92.77$\pm$3.17&1.82$\pm$0.42 \\
\midrule Supervised classifier & \textbf{99.64$\pm$0.00}&\textbf{99.95$\pm$0.00}&\textbf{1.31$\pm$0.01}&\textbf{99.87$\pm$0.01}&\underline{99.75$\pm$0.01}&\textbf{0.47$\pm$0.08} \\
\midrule\rowcolor{lightgreen} S-PUNA w/o C (ours) & 98.97$\pm$0.00&99.86$\pm$0.00&4.74$\pm$0.00&98.36$\pm$0.00&99.57$\pm$0.00&9.38$\pm$0.00 \\
\rowcolor{lightgreen} S-PUNA w/ C (ours) & \underline{99.40$\pm$0.05}&\underline{99.92$\pm$0.01}&\underline{2.23$\pm$0.11}&99.67$\pm$0.04&\textbf{99.91$\pm$0.01}&1.34$\pm$0.15 \\
\midrule \midrule

\multirow{2}{*}{\textbf{Method}} & \multicolumn{6}{c}{\textbf{Average}} \\
\cmidrule(lr){2-7}
 & \multicolumn{2}{c}{\scriptsize AUROC\up} & \multicolumn{2}{c}{\scriptsize AUPR\up} & \multicolumn{2}{c}{\scriptsize FPR95\down} \\
\midrule
KNN \cite{zhang2009reliable} & \multicolumn{2}{c}{64.59$\pm$0.00} & \multicolumn{2}{c}{75.85$\pm$0.00} & \multicolumn{2}{c}{71.73$\pm$0.00} \\
Dist-PU \cite{zhao2022distpu}     & \multicolumn{2}{c}{90.59$\pm$0.25} & \multicolumn{2}{c}{91.92$\pm$0.22} & \multicolumn{2}{c}{26.56$\pm$0.75} \\
saPU  \cite{Dai2025CloserLook}      & \multicolumn{2}{c}{89.01$\pm$0.37} & \multicolumn{2}{c}{86.68$\pm$1.72} & \multicolumn{2}{c}{48.18$\pm$1.67} \\
Dc-PU \cite{Li2025DCPU}  & \multicolumn{2}{c}{65.56$\pm$3.20} & \multicolumn{2}{c}{53.62$\pm$3.02} & \multicolumn{2}{c}{87.59$\pm$5.93} \\
LaGAM \cite{long2024lagam}      & \multicolumn{2}{c}{89.75$\pm$0.33} & \multicolumn{2}{c}{90.38$\pm$1.11} & \multicolumn{2}{c}{41.38$\pm$1.33} \\
\midrule Supervised classifier & \multicolumn{2}{c}{\textbf{98.61$\pm$0.20}} & \multicolumn{2}{c}{\textbf{99.14$\pm$0.15}} & \multicolumn{2}{c}{\textbf{6.34$\pm$1.00}} \\
\midrule\rowcolor{lightgreen} S-PUNA w/o C (ours) & \multicolumn{2}{c}{95.88$\pm$0.00} & \multicolumn{2}{c}{97.79$\pm$0.00} & \multicolumn{2}{c}{19.57$\pm$0.00} \\
\rowcolor{lightgreen} S-PUNA w/ C (ours) & \multicolumn{2}{c}{\underline{98.20$\pm$0.10}} & \multicolumn{2}{c}{\underline{99.00$\pm$0.06}} & \multicolumn{2}{c}{\underline{8.52$\pm$0.48}} \\
\bottomrule
\end{tabular}
\end{table}

\begin{table}[h]
\centering
\vspace{-10pt}
\caption{\textbf{Comparison of methods on \textbf{EuroSAT} with EuroSAT-D (Near shift) and EuroSAT-C (Far shift)}. (\textbf{w/ C}: with classifier; \textbf{w/o C}: without classifier; \textbf{C}: classifier). We report AUROC, AUPR, and FPR95 (\%). Results are shown as \textit{mean} $\pm$ \textit{std}.}
\label{tab:eurosat}
\resizebox{\textwidth}{!}{
\begin{tabular}{l ccc ccc ccc}
\toprule
\multirow{2}{*}{\textbf{Method}} & \multicolumn{3}{c}{\textbf{EuroSAT-D}} & \multicolumn{3}{c}{\textbf{EuroSAT-C}} & \multicolumn{3}{c}{\textbf{Average}} \\
\cmidrule(lr){2-4} \cmidrule(lr){5-7} \cmidrule(lr){8-10}
 & AUROC$\uparrow$ & AUPR$\uparrow$ & FPR95$\downarrow$ & AUROC$\uparrow$ & AUPR$\uparrow$ & FPR95$\downarrow$ & AUROC$\uparrow$ & AUPR$\uparrow$ & FPR95$\downarrow$ \\
\midrule
$k$-NN \cite{zhang2009reliable}     & 58.86$\pm$0.00&84.80$\pm$0.00&13.91$\pm$0.00&81.03$\pm$0.00&83.11$\pm$0.00&58.47$\pm$0.00&69.95$\pm$0.00&83.96$\pm$0.00&36.19$\pm$0.00 \\
Dist-PU \cite{zhao2022distpu}       & 99.78$\pm$0.13 & \textbf{99.95$\pm$0.03} & 0.72$\pm$0.46 &\underline{89.46$\pm$0.50} & \underline{96.48$\pm$0.26} & \underline{49.97$\pm$2.81 }& 94.62$\pm$0.31 & \underline{98.22$\pm$0.14} & 25.35$\pm$1.64\\
saPU \cite{Dai2025CloserLook} & 96.71$\pm$0.93 & 99.30$\pm$0.21 & 7.55$\pm$3.58 & 86.75$\pm$0.62 & 95.81$\pm$0.15 & 54.38$\pm$1.06 & 91.73$\pm$0.78 & 97.56$\pm$0.18 & 30.97$\pm$2.32 \\
Dc-PU \cite{Li2025DCPU} & 68.96$\pm$3.53 & 87.87$\pm$0.59 & 90.89$\pm$6.73 & 60.22$\pm$4.34 & 85.05$\pm$1.43 & 94.48$\pm$3.06 & 64.59$\pm$3.93 & 86.46$\pm$1.01 & 92.68$\pm$4.90 \\
LaGAM \cite{long2024lagam} & 99.33$\pm$0.56 & 99.86$\pm$0.14 & 3.01$\pm$3.39 & 86.24$\pm$0.90 & 96.28$\pm$0.27 & 54.83$\pm$3.24 & 92.79$\pm$0.73 & 98.07$\pm$0.21 & 28.92$\pm$3.32\\
\midrule Supervised classifier  & \textbf{99.91$\pm$0.13}&\underline{98.44$\pm$2.01}&\textbf{0.21$\pm$0.35}&\textbf{100.00$\pm$0.00}&\textbf{100.00$\pm$0.00}&\textbf{0.00$\pm$0.00}&\textbf{99.96$\pm$0.07}&\textbf{99.22$\pm$1.01}&\textbf{0.11$\pm$0.18} \\
\midrule
\rowcolor{lightgreen}\textbf{S-PUNA} w/o C (ours) & 84.40$\pm$0.00&38.02$\pm$0.00&55.76$\pm$0.00&\textbf{100.00$\pm$0.00}&\textbf{100.00$\pm$0.00}&\textbf{0.00$\pm$0.00}&92.20$\pm$0.00&69.01$\pm$0.00&27.88$\pm$0.00 \\
\rowcolor{lightgreen}\textbf{S-PUNA} w/ C (ours) & \underline{99.79$\pm$0.05} & 96.08$\pm$0.70&\underline{0.65$\pm$0.36}&\textbf{100.00$\pm$0.00}&\textbf{100.00$\pm$0.00}&\textbf{0.00$\pm$0.00}&\underline{99.90$\pm$0.03}&98.04$\pm$0.35&\underline{0.33$\pm$0.18} \\

\bottomrule
\end{tabular}
} 
\end{table}

\begin{table}[!h]
\centering
\caption{\textbf{Comparison of methods on \textbf{Tiny-ImageNet} with TinyImageNet-V2 (Near shift) and TinyImageNet-C (Far shift)} (\textbf{w/ C}: with classifier; \textbf{w/o C}: without classifier; \textbf{C}: classifier). We report AUROC, AUPR, and FPR95 (\%). Results are shown as \textit{mean} $\pm$ \textit{std}.}
\label{tab:tinyimagenet}
\resizebox{\textwidth}{!}{
\begin{tabular}{l ccc ccc ccc}
\toprule
\multirow{2}{*}{\textbf{Method}} & \multicolumn{3}{c}{\textbf{TinyImageNet-V2}} & \multicolumn{3}{c}{\textbf{TinyImageNet-C}} & \multicolumn{3}{c}{\textbf{Average}} \\
\cmidrule(lr){2-4} \cmidrule(lr){5-7} \cmidrule(lr){8-10}
 & AUROC$\uparrow$ & AUPR$\uparrow$ & FPR95$\downarrow$ & AUROC$\uparrow$ & AUPR$\uparrow$ & FPR95$\downarrow$ & AUROC$\uparrow$ & AUPR$\uparrow$ & FPR95$\downarrow$ \\
\midrule
$k$-NN \cite{zhang2009reliable}     & 82.00$\pm$0.00&50.65$\pm$0.00&78.39$\pm$0.00&95.82$\pm$0.00&28.19$\pm$0.00&96.96$\pm$0.00&88.91$\pm$0.00&39.42$\pm$0.00&87.68$\pm$0.00 \\
Dist-PU \cite{zhao2022distpu}       & \underline{99.96$\pm$0.00}& \underline{99.98$\pm$0.00} & 0.15$\pm$0.04 & 99.46$\pm$0.11 & 96.14$\pm$0.62 & 2.81$\pm$0.60 & 99.71$\pm$0.06 & 98.06$\pm$0.31 & 1.48$\pm$0.32 \\
saPU \cite{Dai2025CloserLook} & 91.73$\pm$0.01 & 96.37$\pm$0.01 & 19.14$\pm$0.01 & 92.07$\pm$0.17 & 71.47$\pm$1.97 & 33.44$\pm$0.40 & 91.90$\pm$0.09 & 83.92$\pm$0.99 & 26.29$\pm$0.21\\
Dc-PU \cite{Li2025DCPU} & 95.93$\pm$2.90 & 97.82$\pm$1.52 & 16.43$\pm$13.11 & 82.95$\pm$5.58 & 35.21$\pm$14.85 & 61.76$\pm$21.21 & 89.44$\pm$4.24 & 66.52$\pm$8.19 & 39.09$\pm$17.16 \\
LaGAM \cite{long2024lagam}     & 99.93$\pm$0.01 & 99.96$\pm$0.01 & 0.22$\pm$0.01 & 99.59$\pm$0.21 & 95.35$\pm$3.03 & 1.39$\pm$0.78 & 99.76$\pm$0.11 & 97.66$\pm$1.52 & 0.81$\pm$0.39 \\
\midrule Supervised classifier  & \textbf{100.00$\pm$0.01}&\textbf{100.00$\pm$0.00}&\textbf{0.00$\pm$0.00}&\textbf{99.79$\pm$0.23}&\textbf{99.71$\pm$0.32}&\textbf{0.77$\pm$0.98}&\textbf{99.90$\pm$0.12}&\textbf{99.86$\pm$0.16}&\textbf{0.39$\pm$0.49} \\
\midrule
\rowcolor{lightgreen}\textbf{S-PUNA} w/o C (ours) & \textbf{100.00$\pm$0.00}&\textbf{100.00$\pm$0.00}&\textbf{0.00$\pm$0.00}&99.54$\pm$0.00&\underline{99.62$\pm$0.00} & \underline{1.20$\pm$0.00} & \underline{99.77$\pm$0.00}&\underline{99.81$\pm$0.00}&\underline{0.60$\pm$0.00} \\
\rowcolor{lightgreen}\textbf{S-PUNA} w/ C (ours) & 99.90$\pm$0.10&99.92$\pm$0.08&\underline{0.02$\pm$0.04} & \underline{99.64$\pm$0.11} & 99.46$\pm$0.15&1.51$\pm$0.50&\underline{99.77$\pm$0.11}&99.69$\pm$0.12&0.77$\pm$0.27 \\

\bottomrule
\end{tabular}
} 
\end{table}

\section {Ablation Study}\label{appendix:ablation_study_convergence}

Table~\ref{tab:ablation} analyzes the sensitivity of S-PUNA to its main hyperparameters on far shift dataset ImageNet-R and near shift dataset ImageNet-V2. Overall, the results indicate that the proposed method is robust to most hyperparameter choices, with only a few parameters having a significant impact on performance.

\paragraph{Effect of contamination tolerance $\epsilon$.}
The contamination tolerance parameter $\epsilon$ controls how much label noise is tolerated during the pseudo-labeling process. 
Interestingly, the results show that this parameter has virtually no influence on the final performance. 
For $\epsilon \in \{0, 0.05, 0.1\}$, the detection metrics remain identical, reaching $97.34\%$ AUROC, $98.49\%$ AUPR, and $13.37\%$ FPR95 for ImageNet-R and with a similar behavior on ImageNet-V2 . 
This suggests that the contamination score estimated by S-PUNA naturally converges toward small values, making the method insensitive to small variations of $\epsilon$. 
In practice, this indicates that the algorithm is able to control contamination in the pseudo-labeling process without requiring precise tuning of this parameter.

\paragraph{Effect of expansion parameter $\beta$.}
The parameter $\beta$ controls the number of samples that are symmetrically added to the positive and shifted sets during each iteration of the algorithm. 
Unlike $\epsilon$, this parameter has a noticeable impact on the detection performance. 
Increasing $\beta$ progressively improves the results on ImageNet-R, with AUROC increasing from $93.78\%$ for $\beta=500$ to $97.34\%$ for $\beta=1500$, while FPR95 decreases from $31.92\%$ to $13.37\%$. We notice the same trend for ImageNet-V2.
This behavior suggests that larger expansion steps allow the algorithm to more effectively explore the local neighborhood structure of the feature space and progressively uncover the shifted distribution. 
However, the improvement saturates beyond $\beta=1500$, indicating that excessively large expansions do not bring additional benefits.

\paragraph{Effect of neighborhood size $k$.}
The neighborhood size $k$ determines the number of nearest neighbors used to estimate local structure during the manifold expansion stage. 
The results show that S-PUNA remains highly stable across a wide range of values. 
For $k \in \{100, 300, 500\}$, on ImageNet-R AUROC remains close to $97\%$ and FPR95 varies only slightly between $13.37\%$ and $15.50\%$. We observe the same behavior for ImageNet-V2.
This stability suggests that the proposed geometric expansion strategy is robust to moderate variations in the local neighborhood size, which simplifies the practical use of the method.

\paragraph{Effect of initialization size $\alpha$.}
The parameter $\alpha$ defines the number of samples used to initialize the positive and shifted sets before the iterative expansion process begins. 
The results indicate that S-PUNA is stable for moderate initialization sizes ($\alpha=30$ or $50$). 
However, performance degrades significantly when $\alpha$ becomes too large. 
In particular, for ImageNet-R setting $\alpha=100$ leads to a substantial drop in performance, with AUROC decreasing to $72.45\%$ and FPR95 increasing to $89.42\%$. Similarly, for ImageNet-V2, performance drops substantially at this value.
This degradation can be explained by the fact that overly large initial sets introduce noisy samples early in the process, which propagates errors during the neighborhood expansion phase. 
Therefore, smaller initialization sizes provide a cleaner starting point and lead to more reliable expansion dynamics.

\begin{table}[t]
\centering
\caption{\textbf{Ablation study of key hyperparameters of S-PUNA.} When varying one parameter, the others are fixed to their default values. Results are reported on \textbf{ImageNet-R} and \textbf{ImageNet-V2}.}
\label{tab:ablation}
\small
\setlength{\tabcolsep}{3pt}
\resizebox{\linewidth}{!}{%
\begin{tabular}{lccccccc}
\toprule
\multirow{2}{*}{\textbf{Hyperparameter}} 
& \multirow{2}{*}{\textbf{Value}} 
& \multicolumn{3}{c}{\textbf{ImageNet-R}} 
& \multicolumn{3}{c}{\textbf{ImageNet-V2}} \\
\cmidrule(lr){3-5} \cmidrule(lr){6-8}
& 
& \textbf{AUROC}$\uparrow$ 
& \textbf{AUPR}$\uparrow$ 
& \textbf{FPR95}$\downarrow$
& \textbf{AUROC}$\uparrow$ 
& \textbf{AUPR}$\uparrow$ 
& \textbf{FPR95}$\downarrow$ \\
\midrule

\multirow{3}{*}{$\epsilon$}
& 0    & 97.34 & 98.49 & 13.37 & 96.34 &97.69 & 16.93 \\
& 0.05 & 97.34 & 98.49 & 13.37 & 96.34 &97.69 & 16.93 \\
& 0.1  & 97.34 & 98.49 & 13.37 & 96.34 &97.69 & 16.93 \\

\midrule
\multirow{3}{*}{$\beta$}
& 500  & 93.78 & 96.15 & 31.92 & 95.17 &96.59 &24.60 \\
& 1000 & 96.47 & 97.93 & 18.28 & 96.16 & 97.49 & 18.38 \\
& 1500 & 97.34 & 98.49 & 13.37 & 96.34 &97.69 & 16.93 \\

\midrule
\multirow{3}{*}{$k$ (KNN)}
& 100 & 97.34 & 98.49 & 13.37 & 96.34 &97.69 & 16.93  \\
& 300 & 96.98 & 98.22 & 15.50 & 96.07& 97.44 &18.54 \\
& 500 & 97.14 & 98.35 & 14.51 & 96.57 & 97.78 & 14.97 \\

\midrule
\multirow{3}{*}{$\alpha$}
& 30  & 97.34 & 98.49 & 13.37 & 96.34 &97.69 & 16.93 \\
& 50  & 97.23 & 98.42 & 14.22 & 96.36 & 97.61 & 17.79 \\
& 100 & 72.45 & 78.22 & 89.42 & 50.04 & 60.84 & 97.10 \\

\bottomrule
\end{tabular}%
}
\end{table}

\section{Details on ViT layer selection and covariate shift}\label{appendix:vit_layer_selection_covariate_shifts}

To determine which representation layer of the Vision Transformer (ViT-B/16) provides the most suitable features for covariate shift detection, we evaluate the CLS token extracted from each transformer block. 
The analysis relies on three complementary clustering metrics: the Silhouette score (higher is better), the Davies--Bouldin index (lower is better), and the Calinski--Harabasz index (higher is better).

For each experiment, we randomly sample 10,000 images from the ImageNet validation set and 10,000 images from either a covariate shift dataset (ImageNet-C, ImageNet-R, ImageNet-V2, GenImage) or a semantic shift dataset (OOD). 
We then compute clustering scores by comparing the feature representations of these two groups across all transformer blocks.

\begin{figure}[t]
\centering
\includegraphics[width=1.05\linewidth]{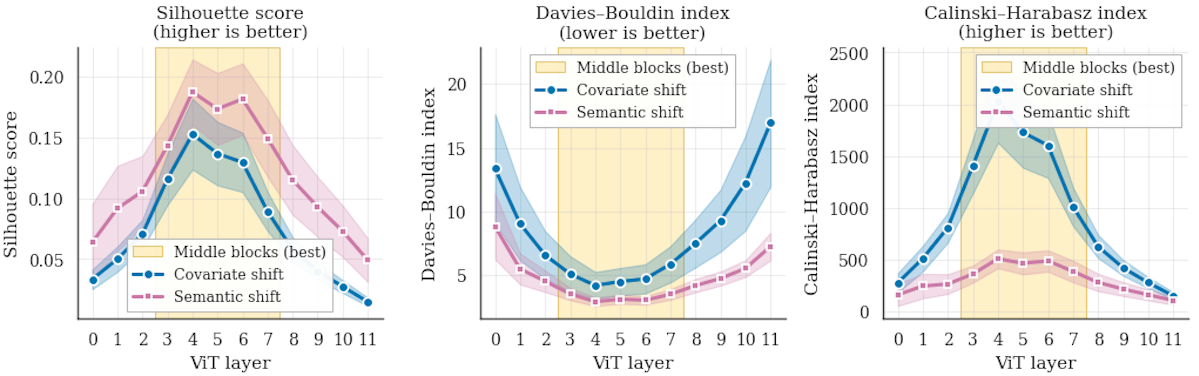}
\caption{\textbf{Layer-wise evaluation of ViT-B/16 representations for distribution shift detection.} 
We assess each transformer block using three clustering quality metrics: the \textit{Silhouette score} (higher values indicate better separation between clusters), the \textit{Davies--Bouldin index} (lower values indicate better cluster separation and compactness), and the \textit{Calinski--Harabasz index} (higher values correspond to better defined and more separated clusters).}
\label{fig:vit_layer_selection}
\label{fig:my_figure}
\end{figure}

For clarity, we group the transformer layers into three regions: \textit{early blocks} (before block 4), \textit{middle blocks} (between blocks 4 and 7), and \textit{late blocks} (after block 7). 
Figure~\ref{fig:vit_layer_selection} shows that the best separation between distributions consistently occurs in the middle layers of the transformer.

Early layers mainly capture low-level visual features such as textures, colors, or simple edges. 
As a result, they are not well suited for detecting distribution shifts that occur at a higher semantic level, since many datasets share similar low-level statistics. On the other hand, the last layers of the transformer become highly specialized for the classification task. These representations strongly focus on semantic class information learned during training and may ignore subtle variations in the input distribution. 
Consequently, they are less effective for detecting covariate shifts that modify the input statistics without changing the semantic labels. 
The middle layers provide a better compromise: they capture intermediate-level structures that still preserve information about the input distribution while remaining sufficiently discriminative. Hence, across the three clustering metrics, the middle blocks (approximately layers 3--7) consistently obtain the most favorable scores. 
Based on this observation, we select block 6 for all experiments, as it provides a strong trade-off between distribution separability and representation stability. We performed similar experiments on ResNet-50 to select the correct representation.

Another important observation is that semantic shift datasets exhibit stronger separation from ImageNet than covariate shift datasets across most layers. 
This behavior is expected because semantic shifts introduce larger distribution changes, typically corresponding to different object categories. 
In contrast, covariate shifts preserve the class semantics while modifying only the input statistics, which makes them more difficult to detect in the feature space.

\section{Generalization of results}\label{appendix:gen}

We evaluate how well a detector trained on one type of distribution shift generalizes to other shifts. Table~\ref{tab:cross_shift} reports AUROC and FPR95 when training on a single shifted dataset and testing on different ones.

Training on \textbf{ImageNetV2} shows strong transfer to synthetic shifts, achieving 91.04 AUROC on GenImage with a very low 0.36 FPR95. However, performance drops on corruption-based shifts such as ImageNet-C (70.01 AUROC, 69.29 FPR95).

Similarly, training on \textbf{ImageNet-R} generalizes reasonably well to GenImage (87.33 AUROC), but remains weaker on ImageNet-C (68.16 AUROC).

Detectors trained on \textbf{ImageNet-C} exhibit poor cross-shift generalization overall, with AUROC dropping to 56.80 on GenImage and 62.79 on ImageNetV2, suggesting that corruption-based shifts do not transfer well to other types of distribution shift.

Finally, training on \textbf{GenImage} transfers well to other distribution shifts, achieving 89.69 AUROC on ImageNetV2 and 81.72 AUROC on ImageNet-R, while performance remains moderate on ImageNet-C (72.57 AUROC).

Overall, GenImage achieves the strongest cross-shift generalization, with the highest average AUROC of 81.33, followed by ImageNetV2 (80.69) and ImageNet-R (78.22). In contrast, detectors trained on ImageNet-C perform substantially worse, with an average AUROC of only 63.59 and the highest average FPR95 of 82.66.

\begin{table}[t]
\centering
\caption{\textbf{Cross-dataset generalization of S-PUNA.} 
Rows indicate the dataset used for PU learning, while columns indicate the dataset used for testing. 
We report AUROC $\uparrow$ and FPR95 $\downarrow$.}
\label{tab:cross_shift}
\scriptsize
\setlength{\tabcolsep}{2pt}

\resizebox{\linewidth}{!}{
\begin{tabular}{lcccccccccc}
\toprule
& \multicolumn{10}{c}{\textbf{Tested on}} \\
\cmidrule(lr){2-11}

\textbf{Trained on}
& \multicolumn{2}{c}{ImageNet-V2}
& \multicolumn{2}{c}{ImageNet-R}
& \multicolumn{2}{c}{GenImage}
& \multicolumn{2}{c}{ImageNet-C}
& \multicolumn{2}{c}{Average} \\

\cmidrule(lr){2-3} \cmidrule(lr){4-5} \cmidrule(lr){6-7} \cmidrule(lr){8-9} \cmidrule(lr){10-11}

& AUROC$\uparrow$ & FPR95$\downarrow$
& AUROC$\uparrow$ & FPR95$\downarrow$
& AUROC$\uparrow$ & FPR95$\downarrow$
& AUROC$\uparrow$ & FPR95$\downarrow$
& AUROC$\uparrow$ & FPR95$\downarrow$ \\

\midrule

ImageNetV2
& -- & --
& 81.03 & 77.80 
& 91.04 & 0.36 
& 70.01 & 69.29
& 80.69 & 49.15 \\

ImageNet-R
& 79.17 & 69.46 
& -- & --
& 87.33 & 40.33
& 68.16 & 67.04
& 78.22 & 58.94 \\

ImageNet-C
& 62.79 & 85.43
& 71.18 & 81.50
& 56.80 & 81.05
& -- & --
& 63.59 & 82.66 \\

GenImage
& 89.69 & 47.64
& 81.72 & 77.70
& -- & --
& 72.57 & 64.75
& 81.33 & 63.36 \\

\bottomrule
\end{tabular}
}

\end{table}

\section{Unsupervised covariate shift detection}\label{appendix:ood_detection_covariateshift}

To better understand how difficult it is to detect covariate shift in large-scale vision models, we evaluate several existing \textbf{unsupervised shift detection approaches}. 
The objective of this section is to study whether covariate shift can be reliably detected in a fully unsupervised setting, that is, without having access to labeled shifted data.

To design our evaluation protocol, we rely on the literature on \textbf{generalized out-of-distribution (OOD) detection}~\cite{yang2024generalized,miyai2025generalized}. 
This line of work studies how models behave when the test distribution differs from the training distribution and proposes a taxonomy of distribution shifts depending on where the change occurs in the joint distribution $P(X,Y)$.

In practice, two main types of shifts are commonly considered. 
\textbf{Semantic shift} corresponds to situations where new classes or new semantic concepts appear at test time. 
In contrast, \textbf{covariate shift} refers to a change in the input distribution while the underlying labeling rule remains the same. 
In other words, the marginal distribution of inputs changes but the relationship between inputs and labels stays constant.

In this section, we focus on \textbf{unsupervised approaches} that do not require any labeled shifted data. 
These approaches can be divided into two main families:

\begin{itemize}

\item \textbf{OOD detection methods}, which compute a score from a trained classifier in order to determine whether a sample deviates from the training distribution.

\item \textbf{Anomaly detection methods}, which attempt to model the distribution of normal data and identify samples that deviate from it.

\end{itemize}

\subsection{OOD detection methods for unsupervised covariate shift detection}

\begin{wrapfigure}[9]{r}{0.65\textwidth}
    \centering
     \vspace{-30pt}
        \includegraphics[width=0.65\textwidth]{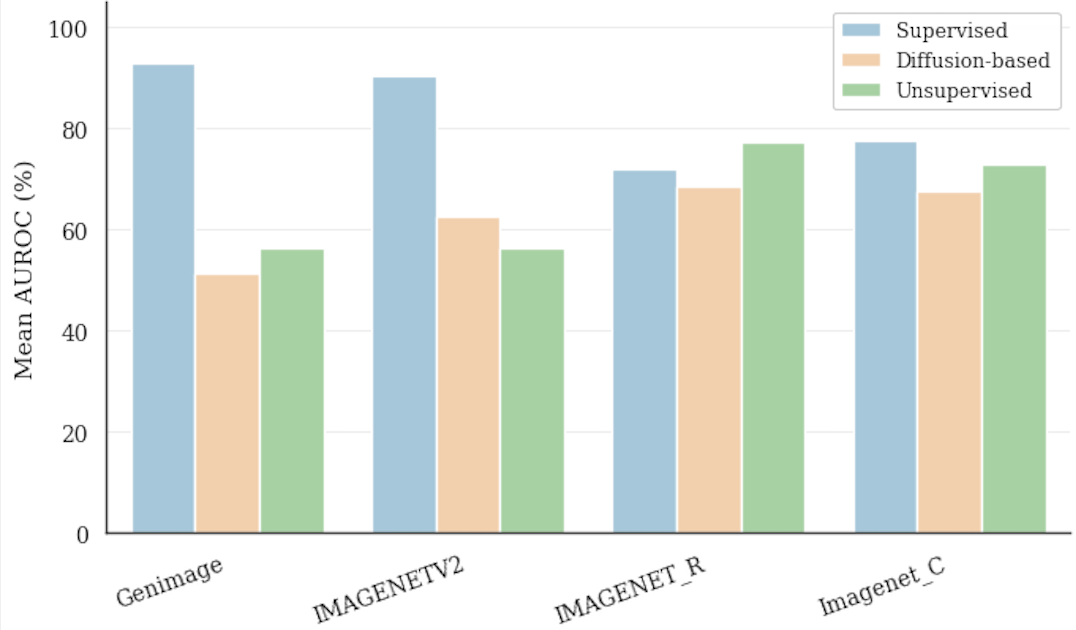}
    \caption{\textbf{Overview of the performance of ood detection methods on covariate shifted datasets of imagenet1k}}
    \label{fig:ood_methods_dataset}
\end{wrapfigure}

OOD detection methods aim to identify samples that do not follow the training distribution by computing a detection score from the outputs or the internal representations of a trained classifier. 
Typical approaches rely on confidence scores, energy-based scores, feature-space distances, or activation manipulation techniques.

These methods can be divided into two categories. 
\textbf{Methods by design} modify the training procedure of the classifier, for example, by introducing additional losses, confidence regularization, or auxiliary datasets containing outliers. 
Although these approaches can improve OOD detection performance, they require retraining the model and often rely on additional data.

In contrast, \textbf{post-hoc methods} operate on a pretrained model without modifying its training procedure. 
They compute an OOD score directly from the model outputs or intermediate features and can therefore be applied in a fully unsupervised way at test time. 
Because they are simple to use and do not require retraining the model, post-hoc methods are widely used in practice.

In our experiments, we focus on \textbf{unsupervised post-hoc OOD detection methods}. 
Most of the evaluated techniques are implemented using the \texttt{OpenOOD} benchmark framework~\cite{yang2022openood}, which provides standardized implementations of many state-of-the-art OOD scoring methods. 
In addition, we also evaluate several recent \textbf{diffusion-based approaches} that compute detection scores using generative models, as well as one supervised baseline for reference.

Table~\ref{tab:ood_results_vit} presents the results obtained on four datasets using ViT-B/16 features. Several observations can be made from these results. First, \textbf{unsupervised post-hoc methods perform poorly on near covariate shifts}. 
For datasets such as ImageNet-V2 and GenImage, which are relatively close to the ImageNet training distribution, most methods obtain AUROC values between $54$ and $59$. 
This performance is close to random detection and indicates that these methods struggle to identify small changes in the input distribution.

Second, performance improves when the shift becomes larger. 
For example, on ImageNet-R, which introduces stronger visual changes through artistic renditions of objects, several feature-based methods achieve significantly higher scores. 
Methods such as \textit{ViM}, \textit{MDS}, and \textit{KNN} reach AUROC values between $85$ and $87$, suggesting that distance-based representations in feature space can capture stronger distribution changes.

Third, the results on ImageNet-C show that corruption-based shifts are moderately detectable. 
Most methods obtain AUROC values around $74$--$76$, which is better than near shifts but still far from perfect detection. 
This suggests that low-level perturbations such as noise, blur, or compression alter the input statistics enough to be partially detected but remain difficult to separate completely.

The results on GenImage highlight another limitation of these approaches. 
Despite the synthetic nature of this dataset, the AUROC values remain close to $55$--$60$ for most methods, showing that generative images can remain close to natural images in feature space and therefore remain difficult to detect.

Overall, these results suggest that \textbf{unsupervised post-hoc OOD detection methods are not well suited for detecting small covariate shifts}. 
They tend to succeed only when the distribution change is sufficiently large.

Finally, the supervised baseline \textit{DisCoPatch} achieves much higher performance on several datasets, for example, reaching AUROC values above $90$ on ImageNet-V2 and GenImage. 
This result highlights the gap between supervised and unsupervised detection methods.

\begin{table*}[t]
\centering
\caption{\textbf{Comparison of OOD detection methods on four shifted datasets using ViT-B/16 features.} 
Methods are grouped by category: unsupervised post-hoc approaches, supervised methods, 
and diffusion/generative-based approaches. We report AUROC $\uparrow$, AUPR $\uparrow$, and FPR95 $\downarrow$.}
\label{tab:ood_results_vit}

\resizebox{\textwidth}{!}{
\begin{tabular}{l ccc ccc ccc ccc}
\toprule
\multirow{2}{*}{\textbf{Method}} 
& \multicolumn{3}{c}{\textbf{ImageNet-V2}} 
& \multicolumn{3}{c}{\textbf{ImageNet-R}} 
& \multicolumn{3}{c}{\textbf{ImageNet-C}} 
& \multicolumn{3}{c}{\textbf{GenImage}} \\

\cmidrule(lr){2-4}
\cmidrule(lr){5-7}
\cmidrule(lr){8-10}
\cmidrule(lr){11-13}

& AUROC & AUPR & FPR95
& AUROC & AUPR & FPR95
& AUROC & AUPR & FPR95
& AUROC & AUPR & FPR95 \\

\midrule
\multicolumn{13}{c}{\textbf{Unsupervised Methods}} \\
\midrule

React\cite{sun2021react} & 56.73 & 22.70 & 93.43 & 81.08 & 76.09 & 74.90 & 75.83 & 76.30 & 69.39 & 55.92 & 15.81 & 93.50 \\
ASH \cite{djurisicextremely} & 54.30 & 21.62 & 94.44 & 72.53 & 70.71 & 92.33 & 50.07 & 55.47 & 96.43 & 50.39 & 12.34 & 95.08 \\
MLS \cite{hendrycks2022scaling} & 56.01 & 22.58 & 94.18 & 74.83 & 71.70 & 92.03 & 74.89 & 76.48 & 75.07 & 56.20 & 16.30 & 93.17 \\
MSP \cite{hendrycks2017a} & 57.51 & 23.05 & 93.16 & 66.95 & 72.24 & 99.86 & 74.73 & 75.47 & 70.71 & 59.15 & 17.29 & 92.22 \\
TempScale \cite{guo2017calibration} & 57.38 & 23.08 & 93.32 & 77.43 & 72.12 & 87.49 & 75.23 & 76.09 & 71.11 & 58.79 & 17.23 & 92.39 \\
Scale \cite{xu2024scaling} & 54.3 & 21.62 & 94.44 & 72.53 & 70.71 & 92.33 & 72.15 & 74.93 & 78.11 & 51.37 & 13.98 & 93.73 \\
VRA \cite{xu2023vra} & 57.29 & 22.95 & 93.47 & 84.52 & 79.03 & 61.65 & 74.84 & 75.22 & 69.60 & 56.68 & 16.18 & 93.30 \\
EBO \cite{liu2020energy} & 54.30 & 21.62 & 94.44 & 72.53 & 70.71 & 92.33 & 73.74 & 86.43 & 87.67 & 53.19 & 14.88 & 93.46 \\
GODIN \cite{hsu2020generalized} & 53.62 & 19.98 & 94.50 & 63.76 & 53.53 & 92.84 & 68.83 & 81.28 & 86.96 & 56.19 & 14.20 & 92.14 \\
RMDS \cite{ren2021simple} & 56.85 & 20.87 & 93.04 & 73.10 & 56.15 & 61.99 & 75.79 & 75.82 & 66.42 & 60.54 & 18.34 & 90.92 \\
MDS \cite{lee2018simple} & 58.55 & 23.94 & 93.44 & 85.95 & 78.65 & 58.09 & 75.11 & 86.08 & 72.97 & 59.41 & 17.99 & 91.34 \\
ViM \cite{wang2022vim} & 57.87 & 23.37 & 93.30 & 87.06 & 81.45 & 56.81 & 76.34 & 87.44 & 77.96 & 56.01 & 17.16 & 94.78 \\
GEN \cite{liu2023gen} & 58.00 & 23.55 & 92.43 & 82.74 & 77.48 & 72.74 & 79.65 & 88.96 & 71.25 & 57.70 & 17.04 & 91.94 \\
KNN \cite{sun2022out} & 58.32 & 23.68 & 93.24 & 85.95 & 80.54 & 58.63 & 75.01 & 86.11 & 72.50 & 59.50 & 18.13 & 91.50 \\

\midrule
\multicolumn{13}{c}{\textbf{Supervised Method}} \\
\midrule

DisCoPatch \cite{caetano2025discopatch} & 90.60 & 73.20 & 47.20 & 71.98 & 64.96 & 88.32 & 77.70 & 71.12 & 45.92 & 92.90 & 69.73 & 33.23 \\

\midrule
\multicolumn{13}{c}{\textbf{Diffusion / Generative Methods}} \\
\midrule

DiffPath\_6D \cite{heng2024out} & 62.40 & 62.50 & 70.67 & 70.27 & 75.67 & 69.83 & 75.72 & 19.17 & 59.65 & 52.60 & 89.66 & 91.52 \\
Lift Map Det -- SIMCLR \cite{liu2023unsupervised} & 62.09 & 62.65 & 71.19 & 72.00 & 49.02 & 98.97 & 73.74 & 46.61 & 74.69 & 50.28 & 88.21 & 94.80 \\
DDPM-OOD \cite{graham2023denoising} & 62.97 & 12.30 & 71.30 & 63.00 & 31.53 & 99.86 & 53.21 & 73.31 & 93.89 & 51.25 & 10.77 & 92.85 \\

\bottomrule
\end{tabular}
}
\end{table*}

\subsection{Anomaly detection methods for unsupervised covariate shift detection}

Anomaly detection methods are another family of techniques that can potentially detect covariate shift. 
As discussed in the generalized OOD detection literature~\cite{yang2024generalized,miyai2025generalized}, anomaly detection typically assumes a setting where only normal data are available during training, and the objective is to detect samples that deviate from this notion of normality at test time. Since covariate shift corresponds to a change in the input distribution, anomaly detection techniques are natural candidates for identifying such changes. In our evaluation we consider a broad set of anomaly detection methods.

\textit{Classical one-class and density-based methods} such as OneClassSVM~\cite{scholkopf2001estimating}, LOF~\cite{breunig2000lof}, ABOD~\cite{kriegel2008angle}, ECOD~\cite{li2022ecod}, and COPOD~\cite{li2020copod} estimate normality from the geometric or statistical structure of the feature space. \textit{Tree-based methods} such as IsolationForest~\cite{liu2008isolation} detect anomalies using recursive partitioning and path-length statistics. \textit{Deep-feature statistical models} such as DFM~\cite{ahuja2019probabilistic} and PaDiM~\cite{defard2021padim} first extract deep features and then model their distribution in order to identify deviations. \textit{Memory-based and flow-based deep anomaly detectors} such as PatchCore~\cite{roth2022towards}, FastFlow~\cite{yu2021fastflow}, and CFA~\cite{lee2022cfa} rely on nearest-neighbor matching, likelihood estimation, or feature transformation in order to capture more complex shifts. Finally, \textit{SimpleNet}~\cite{liu2023simplenet} learns a lightweight discriminator that separates normal features from noise-augmented features and uses the resulting score as an anomaly signal.

Table~\ref{tab:anomaly_results_reordered} presents the performance of these methods. In all methods, we use a ViT-B/16 as encoder for the feature extraction. Overall, the results show that classical anomaly detection methods also struggle to detect near covariate shifts. 
On ImageNet-V2, most methods obtain AUROC values close to $50$, which corresponds to random performance. 
This indicates that small changes in the input distribution are extremely difficult to detect using standard anomaly detection techniques.

Performance improves slightly for stronger shifts. 
For example, on ImageNet-C and ImageNet-R several methods achieve AUROC values above $80$. 
In particular, \textit{PaDiM}, \textit{DFM}, and \textit{CFA} perform relatively well on these datasets, reaching AUROC values around $84$--$86$ on ImageNet-C. 
This suggests that modeling the distribution of deep features can help capture larger changes in the input statistics.

However, performance remains inconsistent across datasets. 
For example, many methods perform reasonably well on ImageNet-C but fail on GenImage, where AUROC values remain around $55$--$62$. 
This again highlights the difficulty of detecting subtle shifts when the feature distributions remain close.

In summary, both OOD detection methods and anomaly detection methods show limited ability to detect covariate shift in a fully unsupervised setting. 
Their performance improves only when the shift becomes large, while near shifts remain extremely difficult to identify.

\begin{table}[t]
\centering
\caption{\textbf{Comparison of anomaly detection methods on four shifted datasets using ViT-B/16 features.} Methods are grouped by family. We report AUROC, AUPR, and FPR95 (\%).}
\label{tab:anomaly_results_reordered}
\resizebox{\textwidth}{!}{
\begin{tabular}{l ccc ccc ccc ccc}
\toprule
\multirow{2}{*}{\textbf{Method}} 
& \multicolumn{3}{c}{\textbf{ImageNet-V2}} 
& \multicolumn{3}{c}{\textbf{ImageNet-R}} 
& \multicolumn{3}{c}{\textbf{ImageNet-C}} 
& \multicolumn{3}{c}{\textbf{GenImage}} \\
\cmidrule(lr){2-4} \cmidrule(lr){5-7} \cmidrule(lr){8-10} \cmidrule(lr){11-13}

& AUROC$\uparrow$ & AUPR$\uparrow$ & FPR95$\downarrow$
& AUROC$\uparrow$ & AUPR$\uparrow$ & FPR95$\downarrow$
& AUROC$\uparrow$ & AUPR$\uparrow$ & FPR95$\downarrow$
& AUROC$\uparrow$ & AUPR$\uparrow$ & FPR95$\downarrow$ \\
\midrule

\multicolumn{13}{c}{\textbf{Classical one-class / density-based methods}} \\
\midrule

OneClassSVM \cite{scholkopf2001estimating} & 50.31 & 37.88 & 95.23 & 53.92 & 36.05 & 97.96 & 73.29 & 67.58 & 61.33 & 57.46 & 12.68 & 90.17 \\
LOF \cite{breunig2000lof} & 50.61 & 38.41 & 94.31 & 55.47 & 44.70 & 94.99 & 84.29 & 83.47 & 42.48 & 61.59 & 19.40 & 84.13 \\
ABOD \cite{kriegel2008angle} & 50.84 & 38.03 & 94.36 & 52.19 & 39.70 & 95.61 & 70.21 & 68.61 & 77.95 & 55.26 & 13.40 & 92.35 \\
ECOD \cite{li2022ecod} & 52.98 & 35.75 & 95.75 & 52.23 & 39.21 & 95.02 & 59.94 & 46.20 & 93.67 & 60.14 & 10.19 & 90.59 \\
COPOD \cite{li2020copod} & 54.04 & 35.10 & 95.97 & 57.32 & 41.93 & 91.14 & 74.21 & 36.95 & 98.89 & 59.15 & 10.20 & 87.45 \\

\midrule
\multicolumn{13}{c}{\textbf{Tree-based methods}} \\
\midrule

IsolationForest \cite{liu2008isolation} & 55.20 & 35.35 & 97.39 & 50.72 & 42.73 & 98.37 & 74.93 & 71.64 & 55.36 & 61.28 & 13.07 & 90.22 \\

\midrule
\multicolumn{13}{c}{\textbf{Deep-feature statistical models}} \\
\midrule

DFM \cite{ahuja2019probabilistic} & 52.36 & 36.21 & 96.26 & 60.70 & 52.60 & 94.07 & 84.60 & 81.92 & 38.56 & 59.89 & 15.55 & 86.77 \\
Padim \cite{defard2021padim} & 50.28 & 38.54 & 95.16 & 61.54 & 52.54 & 93.79 & 85.57 & 81.53 & 33.73 & 62.36 & 16.23 & 80.94 \\

\midrule
\multicolumn{13}{c}{\textbf{Memory-based / flow-based deep anomaly detectors}} \\
\midrule

Patchcore \cite{roth2022towards} & 50.03 & 37.69 & 94.06 & 53.49 & 41.01 & 95.59 & 74.31 & 74.87 & 76.39 & 56.57 & 15.45 & 91.57 \\
FastFlow \cite{yu2021fastflow} & 51.81 & 36.55 & 95.96 & 50.43 & 39.46 & 95.76 & 74.07 & 72.03 & 66.04 & 55.32 & 10.79 & 94.88 \\
CFA \cite{lee2022cfa} & 55.06 & 34.77 & 97.03 & 63.73 & 55.03 & 91.63 & 83.58 & 81.07 & 41.03 & 59.06 & 15.56 & 87.81 \\

\midrule
\multicolumn{13}{c}{\textbf{Others}} \\
\midrule

SimpleNet \cite{liu2023simplenet} & 53.17 & 15.37 & 95.91 & 58.19 & 33.11 & 98.18 & 59.89 & 45.72 & 95.96 & 59.61 & 44.08 & 98.35 \\

\bottomrule
\end{tabular}
}
\end{table}

\section{ResNet Results}\label{appendix:resnet}

Table~\ref{tab:Resnet50} reports the performance of \textbf{S-PUNA} using ResNet50 features across four ImageNet shifted datasets. We compare our approach with the weakly supervised positive-unlabeled (PU) learning baselines, and for reference, we also report a fully supervised classifier, which represents an upper bound since it relies on labeled supervision unavailable to PU methods.

Across all datasets, \textbf{S-PUNA} substantially outperforms existing PU approaches.
On ImageNet-V2, the best PU baseline (\textbf{Dc-PU}) reaches $55.72\%$ AUROC, whereas \textbf{S-PUNA} achieves $84.64\%$ AUROC together with a large reduction of FPR95 to $64.26\%$. This corresponds to an improvement of $+28.92$ over the strongest PU baseline. On ImageNet-R the strongest PU baseline (\textbf{LaGAM}) obtains $80.92\%$ AUROC while \textbf{S-PUNA} improves performance to $88.76\%$ AUROC and reduces FPR95 from $82.03\%$ to $56.08\%$ corresponding to a gain of $+25.95$. On GenImage \textbf{S-PUNA} also provides significant gains. While \textbf{LaGAM} reaches $80.12\%$ AUROC, \textbf{S-PUNA} increases performance to $93.72\%$ AUROC and lowers FPR95 to $31.08\%$, yielding a large improvement of $+13.60$ on AUROC and $+42.48$ on FPR95. Finally, on ImageNet-C, \textbf{S-PUNA} nearly matches the supervised upper bound. It achieves $99.34\%$ AUROC with $2.59\%$ FPR95, outperforming all PU baselines and approaching the supervised classifier ($99.36\%$ AUROC).

Overall, \textbf{S-PUNA} consistently delivers the strongest performance among PU methods. Averaged across datasets, it reaches $91.62\%$ AUROC compared to $77.48\%$ for the best competing PU baseline (\textbf{LaGAM}), corresponding to an improvement of $+14.14$ while also reducing the average FPR95 from $67.87\%$ to $38.50\%$. Although the supervised classifier remains an upper bound ($93.55\%$ AUROC), \textbf{S-PUNA} significantly narrows the gap while relying only on weak supervision.

\begin{table}[t]
\centering
\caption{\textbf{Main Results: ResNet50 Performance comparison across four ImageNet variants and Overall Average}(\textbf{w/ C}: with classifier; \textbf{w/o C}: without classifier; \textbf{C}: classifier). Results are shown as \textit{mean} $\pm$ \textit{std}.}
\label{tab:Resnet50}
\scriptsize
\setlength{\tabcolsep}{2pt} 
\begin{tabular}{l ccc ccc}
\toprule
\multirow{2}{*}{\textbf{Method}} & \multicolumn{3}{c}{\textbf{ImageNet-V2}} & \multicolumn{3}{c}{\textbf{ImageNet-R}} \\
\cmidrule(lr){2-4} \cmidrule(lr){5-7}
 & \scriptsize AUROC\up & \scriptsize AUPR\up & \scriptsize FPR95\down & \scriptsize AUROC\up & \scriptsize AUPR\up & \scriptsize FPR95\down \\
\midrule
KNN \cite{zhang2009reliable} & 50.28$\pm$0.00&65.75$\pm$0.00&94.43$\pm$0.00&69.88$\pm$0.00&77.43$\pm$0.00&88.63$\pm$0.00 \\
Dist-PU \cite{zhao2022distpu}     & 50.65$\pm$0.11&65.63$\pm$0.06&94.69$\pm$0.11&74.54$\pm$0.15&80.36$\pm$0.16&86.46$\pm$0.47 \\
saPU  \cite{Dai2025CloserLook}      & 54.30$\pm$0.83&68.75$\pm$0.71&92.32$\pm$0.58&68.82$\pm$0.59&73.16$\pm$0.07&95.06$\pm$0.50 \\
Dc-PU \cite{Li2025DCPU} & 55.72$\pm$4.60&68.86$\pm$3.65&93.32$\pm$2.59&70.99$\pm$1.59&78.02$\pm$1.61&88.22$\pm$2.51 \\
LaGAM  \cite{long2024lagam}      & 53.80$\pm$10.21&71.43$\pm$1.19&91.89$\pm$1.10&80.92$\pm$1.63&84.35$\pm$1.75&82.03$\pm$3.88 \\
\midrule Supervised classifier & \textbf{88.10$\pm$0.32}&\textbf{92.07$\pm$0.26}&\textbf{51.56$\pm$1.69}&\textbf{91.17$\pm$0.20}&\textbf{94.09$\pm$0.16}&\textbf{43.71$\pm$0.79} \\
\midrule\rowcolor{lightgreen} S-PUNA w/o C (ours) & 70.22$\pm$0.00&80.54$\pm$0.00&80.64$\pm$0.00&86.21$\pm$0.00&90.87$\pm$0.00&59.57$\pm$0.00 \\
\rowcolor{lightgreen} S-PUNA w/ C (ours) & \underline{84.64$\pm$0.27}&\underline{89.21$\pm$0.26}&\underline{64.26$\pm$1.02}&\underline{88.76$\pm$0.20}&\underline{91.90$\pm$0.21}&\underline{56.08$\pm$0.80} \\
\midrule \midrule 

\multirow{2}{*}{\textbf{Method}} & \multicolumn{3}{c}{\textbf{GenImage}} & \multicolumn{3}{c}{\textbf{ImageNet-C}} \\
\cmidrule(lr){2-4} \cmidrule(lr){5-7}
 & \scriptsize AUROC\up & \scriptsize AUPR\up & \scriptsize FPR95\down & \scriptsize AUROC\up & \scriptsize AUPR\up & \scriptsize FPR95\down \\
\midrule
KNN  \cite{zhang2009reliable}  & 67.33$\pm$0.00&66.65$\pm$0.00&82.82$\pm$0.00&78.11$\pm$0.00&77.55$\pm$0.00&52.29$\pm$0.00 \\
Dist-PU \cite{zhao2022distpu}      & 61.55$\pm$0.56&55.99$\pm$0.57&95.15$\pm$0.41&93.29$\pm$0.50&88.63$\pm$0.98&30.31$\pm$2.54 \\
saPU  \cite{Dai2025CloserLook}       & 67.01$\pm$2.38&57.56$\pm$1.65&96.53$\pm$0.15&93.48$\pm$0.67&84.86$\pm$2.34&38.11$\pm$6.89 \\
Dc-PU \cite{Li2025DCPU}  & 60.61$\pm$7.57&57.77$\pm$5.90&92.88$\pm$6.20&75.74$\pm$4.80&58.96$\pm$6.71&72.30$\pm$7.61 \\
LaGAM  \cite{long2024lagam}      & 80.12$\pm$0.81&77.45$\pm$1.20&73.56$\pm$2.23&95.07$\pm$0.67&89.85$\pm$1.88&24.01$\pm$3.96 \\
\midrule Supervised classifier & \textbf{95.57$\pm$0.34}&\textbf{99.18$\pm$0.07}&\textbf{21.13$\pm$1.97}&\textbf{99.36$\pm$0.05}&\underline{98.86$\pm$0.10}&\textbf{2.48$\pm$0.36} \\
\midrule\rowcolor{lightgreen} S-PUNA w/o C (ours) & 83.18$\pm$0.00&97.13$\pm$0.00&49.30$\pm$0.00&97.18$\pm$0.00&99.56$\pm$0.00&8.64$\pm$0.00 \\
\rowcolor{lightgreen} S-PUNA w/ C (ours) & \underline{93.72$\pm$0.18}&\underline{98.86$\pm$0.03}&\underline{31.08$\pm$1.16}&\underline{99.34$\pm$0.05}&\textbf{99.89$\pm$0.01}&\underline{2.59$\pm$0.17} \\
\midrule \midrule

\multirow{2}{*}{\textbf{Method}} & \multicolumn{6}{c}{\textbf{Average}} \\
\cmidrule(lr){2-7}
 & \multicolumn{2}{c}{\scriptsize AUROC\up} & \multicolumn{2}{c}{\scriptsize AUPR\up} & \multicolumn{2}{c}{\scriptsize FPR95\down} \\
\midrule
KNN \cite{zhang2009reliable} & \multicolumn{2}{c}{66.40$\pm$0.00} & \multicolumn{2}{c}{71.85$\pm$0.00} & \multicolumn{2}{c}{79.54$\pm$0.00} \\
Dist-PU \cite{zhao2022distpu}     & \multicolumn{2}{c}{70.01$\pm$0.33} & \multicolumn{2}{c}{72.65$\pm$0.44} & \multicolumn{2}{c}{76.65$\pm$0.88} \\
saPU  \cite{Dai2025CloserLook}      & \multicolumn{2}{c}{70.90$\pm$1.12} & \multicolumn{2}{c}{71.08$\pm$1.19} & \multicolumn{2}{c}{80.51$\pm$2.03} \\
Dc-PU \cite{Li2025DCPU}  & \multicolumn{2}{c}{65.76$\pm$4.64} & \multicolumn{2}{c}{65.90$\pm$4.47} & \multicolumn{2}{c}{86.68$\pm$4.73} \\
LaGAM \cite{long2024lagam}      & \multicolumn{2}{c}{77.48$\pm$3.33} & \multicolumn{2}{c}{80.77$\pm$1.51} & \multicolumn{2}{c}{67.87$\pm$2.79} \\
\midrule Supervised classifier & \multicolumn{2}{c}{\textbf{93.55$\pm$0.23}} & \multicolumn{2}{c}{\textbf{96.05$\pm$0.15}} & \multicolumn{2}{c}{\textbf{29.72$\pm$1.20}} \\
\midrule\rowcolor{lightgreen} S-PUNA w/o C (ours) & \multicolumn{2}{c}{84.20$\pm$0.00} & \multicolumn{2}{c}{92.03$\pm$0.00} & \multicolumn{2}{c}{49.54$\pm$0.00} \\
\rowcolor{lightgreen} S-PUNA w/ C (ours) & \multicolumn{2}{c}{\underline{91.62$\pm$0.18}} & \multicolumn{2}{c}{\underline{94.97$\pm$0.13}} & \multicolumn{2}{c}{\underline{38.50$\pm$0.79}} \\
\bottomrule
\end{tabular}
\end{table}

\section{Near and Far Covariate Shift}\label{appendix:near_vs_far_covariate_shift}\label{appendix:near_vs_far_covariate}

In real-world deployments, test data often deviates from the training distribution; we refer to this mismatch broadly as \textit{distribution shift}. Following the generalized OOD viewpoint, shifts can be described by where the change is manifested under the joint distribution $P(X,Y)$, namely the input space $X$ versus the label space $Y$ \cite{yang2024generalized}. In our setting, \textbf{Covariate Shift} primarily alters the input marginal $P(X)$ while keeping the label space fixed, whereas \textbf{Semantic Shift} is associated with changes in label semantics or effective class composition, which typically induces changes in $P(X)$~\cite{yang2024generalized}.

\paragraph{Near vs. Far Covariate Shift.}
To further characterize the difficulty of detecting covariate shifts, we distinguish between \textit{near} and \textit{far} covariate shifts according to the distance between the input distributions $P_{\text{ID}}(X)$ and $P_{\text{Shift}}(X)$. 
Let $d(\cdot,\cdot)$ denote a distance between distributions
We define the magnitude of the covariate shift as
\[
\Delta_X = d\!\left(P_{\text{ID}}(X),\, P_{\text{Shift}}(X)\right).
\]

A \textbf{Near Covariate Shift} corresponds to situations where $\Delta_X$ is small, meaning that the shifted samples remain close to the training distribution in feature space. 
These cases typically arise from mild domain perturbations such as small changes in acquisition conditions or dataset re-sampling.

In contrast, a \textbf{Far Covariate Shift} occurs when $\Delta_X$ becomes large, indicating that the shifted distribution lies further away from the training manifold. 
Such shifts often arise from strong corruptions, domain changes, or stylistic transformations that significantly alter the input statistics while preserving the semantic labels.

\paragraph{Empirical analysis of covariate shift magnitude.}
To quantify these effects, Table~\ref{tab:near_far_covshift} reports feature-space distances between ImageNet validation samples and several shifted datasets using embeddings extracted from a ViT backbone. 
Distances are computed using both cosine distance and mean $L_2$ distance at two representation levels: an intermediate layer (\texttt{block6}) and the final layer (\texttt{last}).

Several observations can be drawn from these results. 
First, \textbf{ImageNet-V2} consistently exhibits the smallest distances to the ImageNet validation set, with cosine distance $0.0041$ and $L_2$ distance $0.0793$ at block6. 
This indicates that ImageNet-V2 remains extremely close to the original distribution and therefore constitutes a typical \textit{near covariate shift}. 
Similarly, \textbf{GenImage} also remains relatively close to the ImageNet distribution, with moderate distances at block6 (cosine distance $0.0144$), suggesting another form of near-to-moderate shift.

In contrast, \textbf{ImageNet-R} produces significantly larger feature-space deviations, with cosine distance increasing to $0.0366$ at block6 and reaching $0.3440$ at the final layer. 
This indicates that stylized renditions of ImageNet classes create a much stronger distributional deviation, making ImageNet-R representative of a \textit{far covariate shift}.

The strongest shifts are observed for \textbf{ImageNet-C}, where the distance from the training distribution increases with the corruption severity level. 
For instance, at block6 the cosine distance increases from $0.0710$ at severity $1$ to $0.0735$ at severity $5$, while the $L_2$ distance rises from $0.3309$ to $0.3329$. 
At the final representation layer, this effect becomes even more pronounced, with cosine distance reaching $0.3829$ at severity $5$. 
These results indicate that ImageNet-C constitutes a progressively stronger \textit{far covariate shift} as the corruption severity increases.

Overall, this analysis allows us to categorize the considered datasets according to the magnitude of the covariate shift:
\begin{itemize}
    \item \textbf{Near covariate shifts:} ImageNet-V2 and GenImage.
    \item \textbf{Far covariate shifts:} ImageNet-R and ImageNet-C (all severity levels).
\end{itemize}

This distinction is important because near shifts are typically more difficult to detect: since the shifted samples remain close to the training manifold, classical distributional detectors often fail to identify them reliably. In contrast, far shifts generate stronger deviations in feature space and are therefore easier to detect. Our experiments demonstrate that S-PUNA remains effective in both regimes, including the particularly challenging near-shift scenario.

\begin{table}[!ht]
    \centering
    \small
    \setlength{\tabcolsep}{6pt}
    \caption{\textbf{Feature-space proximity between ImageNet(val) and shifted test distributions under different representation layers}. Cosine distance and $L_2$ mean are computed on ViT mean of the embeddings extracted at \texttt{block6} (intermediate) and \texttt{last} (terminal) layers.\label{tab:near_far_covshift}}
    \resizebox{\textwidth}{!}{

    \begin{tabular}{l l l l r r}
        \toprule
        Source & Target & Detail & Layer  & Cosine dist & $L_2$ mean \\
        \midrule
        ImageNet(val) & ImageNet-V2 & -- & block6 & 0.0041 & 0.0793 \\
        ImageNet(val) & GenImage & -- & block6 & 0.0144 & 0.1476 \\
        ImageNet(val) & ImageNet-R & -- & block6 & 0.0366 & 0.2375 \\
        ImageNet(val) & ImageNet-C & severity=1 & block6 & 0.0710 & 0.3309 \\
        ImageNet(val) & ImageNet-C & severity=2 & block6 & 0.0651 & 0.3155 \\
        ImageNet(val) & ImageNet-C & severity=3 & block6 & 0.0670 & 0.3192 \\
        ImageNet(val) & ImageNet-C & severity=4 & block6 & 0.0694 & 0.3241 \\
        ImageNet(val) & ImageNet-C & severity=5 & block6 & 0.0735 & 0.3329 \\
        \midrule
        ImageNet(val) & ImageNet-V2 & -- & last & 0.0151 & 0.0389 \\
        ImageNet(val) & GenImage & -- & last & 0.0744 & 0.0875 \\
        ImageNet(val) & ImageNet-R & -- & last & 0.3440 & 0.2449 \\
        ImageNet(val) & ImageNet-C & severity=1 & last & 0.2153 & 0.1560 \\
        ImageNet(val) & ImageNet-C & severity=2 & last & 0.2286 & 0.1602 \\
        ImageNet(val) & ImageNet-C & severity=3 & last & 0.2553 & 0.1699 \\
        ImageNet(val) & ImageNet-C & severity=4 & last & 0.2999 & 0.1869 \\
        ImageNet(val) & ImageNet-C & severity=5 & last & 0.3829 & 0.2165 \\
        \bottomrule
    \end{tabular}
    }
\end{table}

\section{Downstream Applications}
\label{appendix:Downstream}

We further evaluate S-PUNA on some downstream applications. We consider three practical scenarios: \textbf{image forgery detection}, \textbf{noisy dataset cleaning}, and \textbf{deployment-shift monitoring}. In all cases, the objective is not to adapt the model to the shifted distribution but rather to identify samples whose input statistics differ from the positive domain. These experiments illustrate how S-PUNA can be used as a practical detection tool when only positive domain data and an unlabeled pool are available.

\subsection{Image forgery detection.}
A first practical use case is image forgery detection. In this setting, the positive domain corresponds to real images, while the shifted distribution corresponds to images generated by generative models. This matches our GenImage~\cite{zhu2023genimage} evaluation, where ImageNet1K~\cite{deng2009imagenet} images are used as the positive domain and GenImage~\cite{zhu2023genimage} as shifted samples. The covariate shift here is induced by the image generation process, although generated images may preserve the same semantic classes as ImageNet1K they often contain subtle statistical artifacts in texture, frequency, or local structure. We apply S-PUNA on top of visual features extracted from a pretrained ViT-B/16 on Imagenet1K. S-PUNA then progressively identifies generated samples in the unlabeled pool by exploiting the local geometry of this feature space, reaching \textbf{93.72\% AUROC}. This experiment shows that our method is a practical tool for detecting synthetic or manipulated visual content.

\subsection{Noisy data cleaning.}
A second downstream application is dataset cleaning. In many practical training pipelines, datasets may contain corrupted, degraded, or distributionally shifted samples. Training directly on such noisy data can reduce downstream classification performance. To simulate this setting, we construct a noisy TinyImageNet~\cite{deng2009imagenet} training set composed of 25\% clean TinyImageNet samples and 75\% corrupted TinyImageNet-C~\cite{hendrycksbenchmarking} samples. Here, clean TinyImageNet represents the positive domain while TinyImageNet-C represents the shifted distribution. The shift is caused by image corruptions such as noise, blur, weather effects, or other perturbations that modify the input statistics while preserving the semantic class labels. For this experiment, the downstream classifier is trained for 2 epochs on the training dataset starting from a ViT-B/16 model pre-trained on ImageNet-1K. When the training dataset is the clean TinyImageNet, the model achieves an accuracy of \textbf{87.5\%}. When training on the noisy mixture, accuracy drops to \textbf{81.43\%}, confirming that the corrupted samples harm downstream learning. We use S-PUNA on the original non-fine-tuned ViT checkpoint to detect and remove corrupted samples before training the downstream classifier, and accuracy recovers to \textbf{85.84\%}. This result shows that S-PUNA can be used as a dataset cleaning tool by identifying corrupted samples before training, which recovers a large part of the performance lost due to noisy data.

\subsection{Deployment shift monitoring.}
A third application is deployment monitoring, where the goal is to detect when incoming data no longer matches the expected operating distribution. This is especially important in safety-critical domains such as autonomous driving, where changes in acquisition conditions can affect model reliability. We evaluate this setting using the ACDC dataset ~\cite{sakaridis2025acdc}  where normal driving images are treated as the positive domain and adverse-condition images are treated as shifted samples. The covariate shift corresponds to changes in visual conditions between normal and adverse driving scenes, such as degraded visibility, illumination changes, or weather-related appearance changes. For feature extraction, we use Dinov2\_vitb14 without any fine-tuning, and S-PUNA is applied on block-6 DINOv2 features. S-PUNA achieves \textbf{82.83\% AUROC}, showing that it can detect deployment changes without requiring fully supervised labels for the adverse conditions. This suggests that S-PUNA can serve as a lightweight monitoring tool for flagging distributional changes during deployment.

\clearpage

\bibliographystyle{splncs04}
\bibliography{main}

\end{document}